\documentclass[10pt,twocolumn,letterpaper]{article}

\usepackage{iccv}              
\definecolor{iccvblue}{rgb}{0.21,0.49,0.74}
\usepackage[pagebackref,breaklinks,colorlinks,allcolors=iccvblue]{hyperref}
\hypersetup{urlcolor = magenta}
\title{Domain Generalizable Portrait Style Transfer}

\author{
\begin{tabular}{@{}c@{\hspace{15pt}}c@{\hspace{15pt}}c@{\hspace{15pt}}c@{}}
Xinbo Wang\textsuperscript{1}\footnotemark[2]\and
Wenju Xu\textsuperscript{2}\footnotemark[2]\and
Qing Zhang\textsuperscript{1,3}\footnotemark[1]\and
Wei-Shi Zheng\textsuperscript{1,3}
\end{tabular}\\
\textsuperscript{1} School of Computer Science and Engineering, Sun Yat-sen University, China\quad 
\textsuperscript{2} AMAZON \\
\textsuperscript{3} Key Laboratory of Machine Intelligence and Advanced Computing, Ministry of Education, China\\
{\tt\small wangxb29@mail2.sysu.edu.cn} \quad 
{\tt\small xuwenju123@gmail.com}\\
{\tt\small zhangq93@mail.sysu.edu.cn} \quad 
{\tt\small wszheng@ieee.org}\\
}
\begin{document}
\twocolumn[{
    \maketitle
    \vspace*{-4mm}
    \begin{center}
    \captionsetup{type=figure}
    \captionsetup[subfigure]{labelformat=empty}
    \begin{minipage}[b]{1.00\textwidth}
    \begin{subfigure}[c]{0.160\textwidth}
        \includegraphics[width=1.0\textwidth]{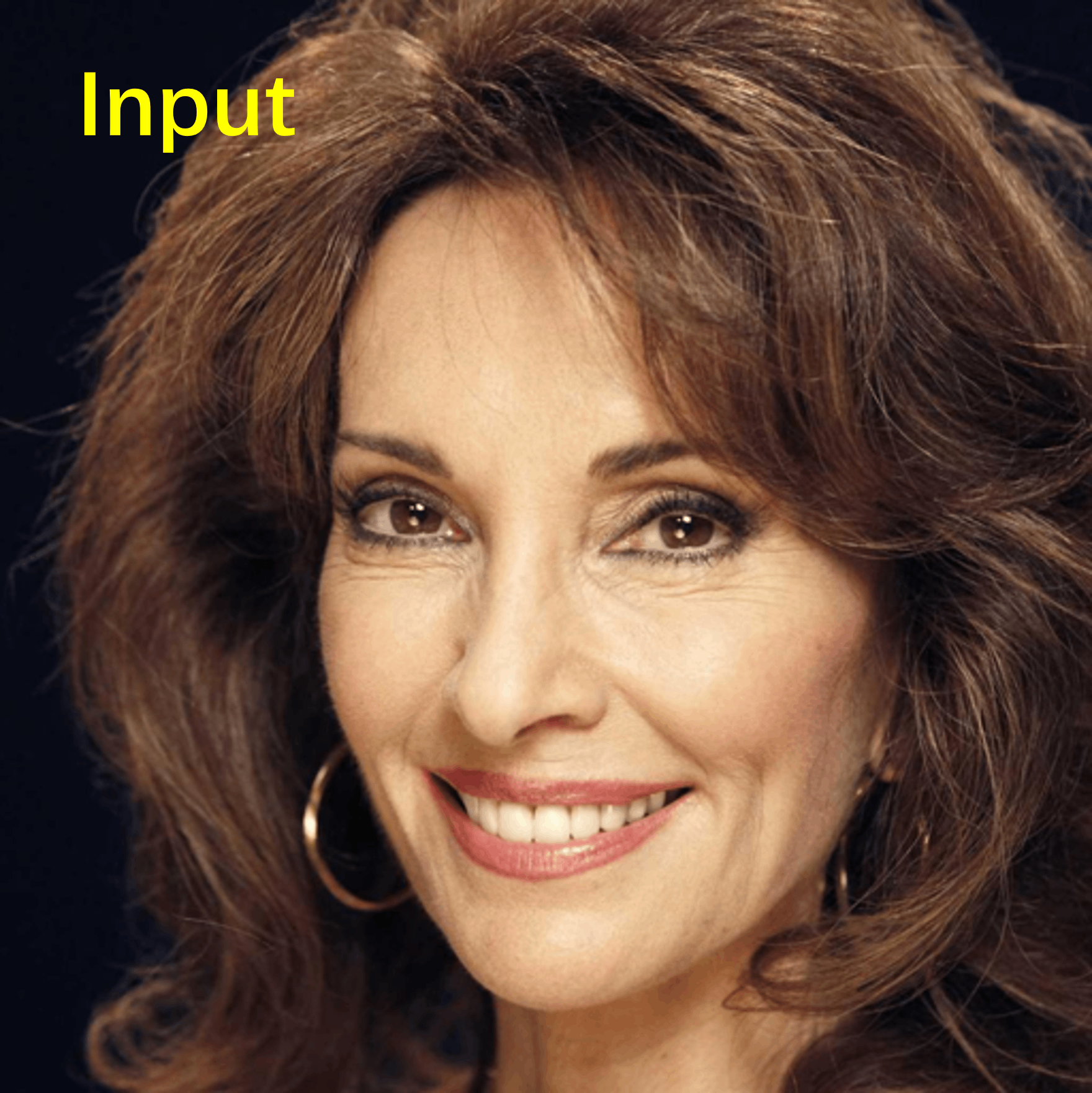}\vspace{1pt}
        \includegraphics[width=1.0\textwidth]{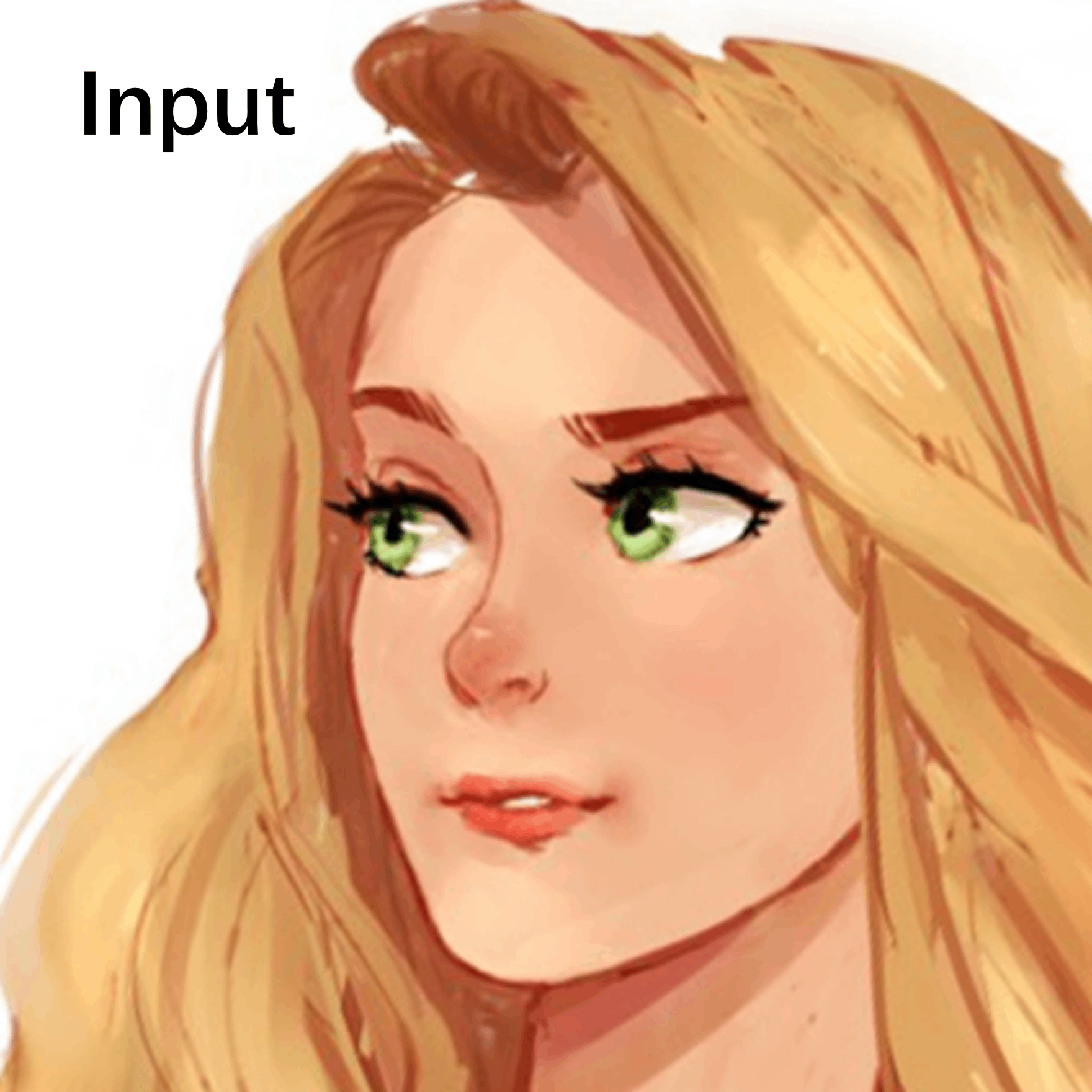}
    \end{subfigure}        
    \begin{subfigure}[c]{0.160\textwidth}
        \includegraphics[width=1.0\textwidth]{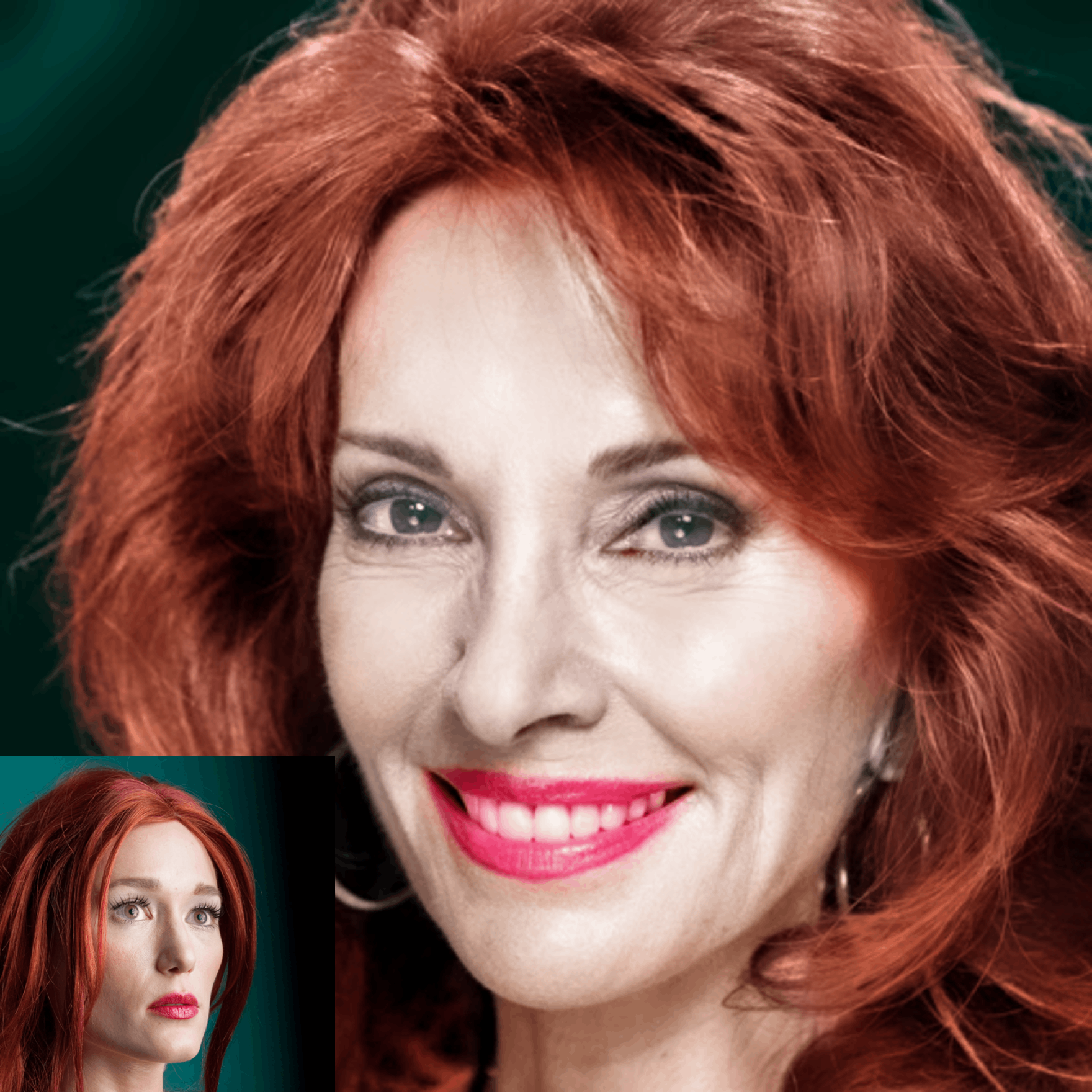}\vspace{1pt}
        \includegraphics[width=1.0\textwidth]{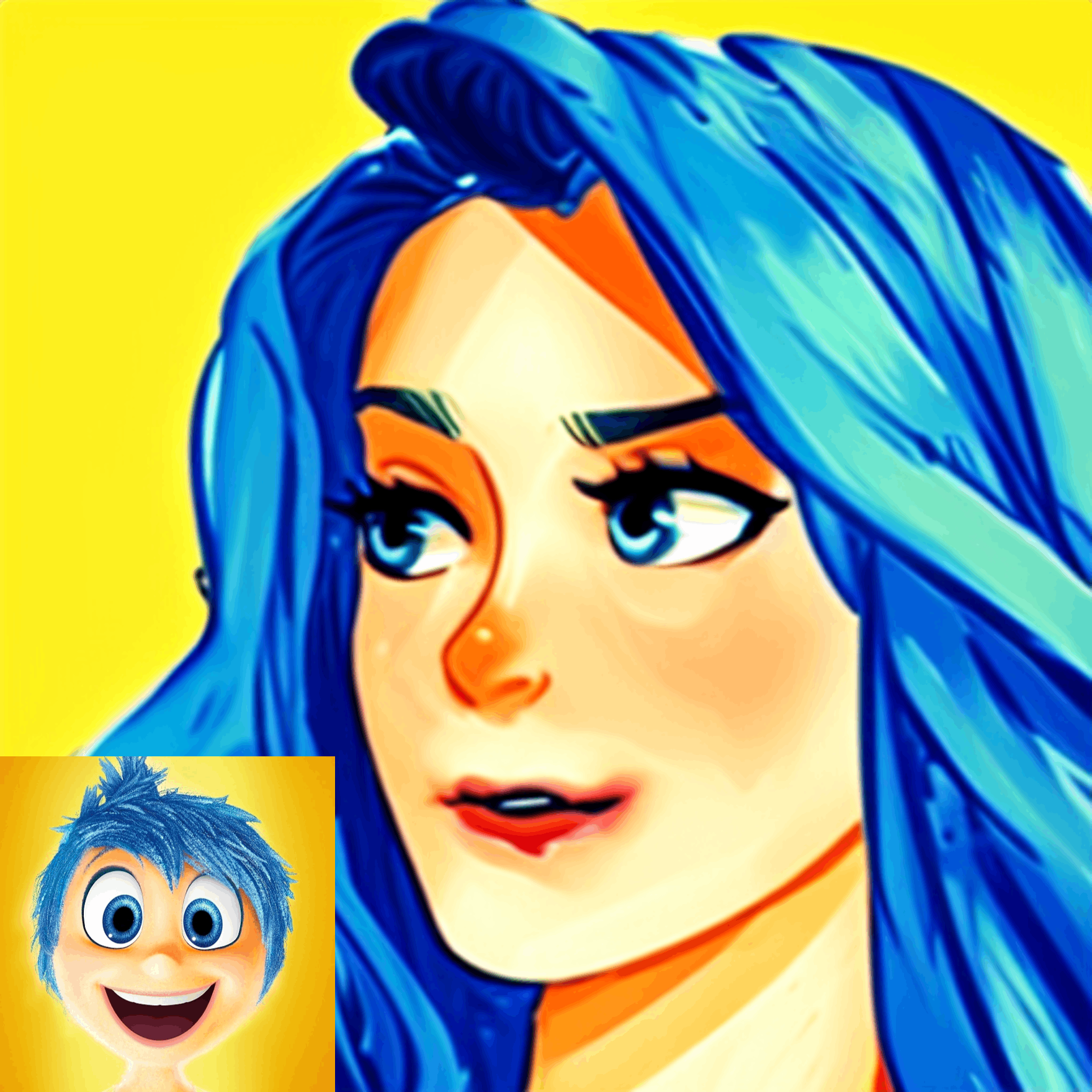}
    \end{subfigure}
    \begin{subfigure}[c]{0.160\textwidth}
        \includegraphics[width=1.0\textwidth]{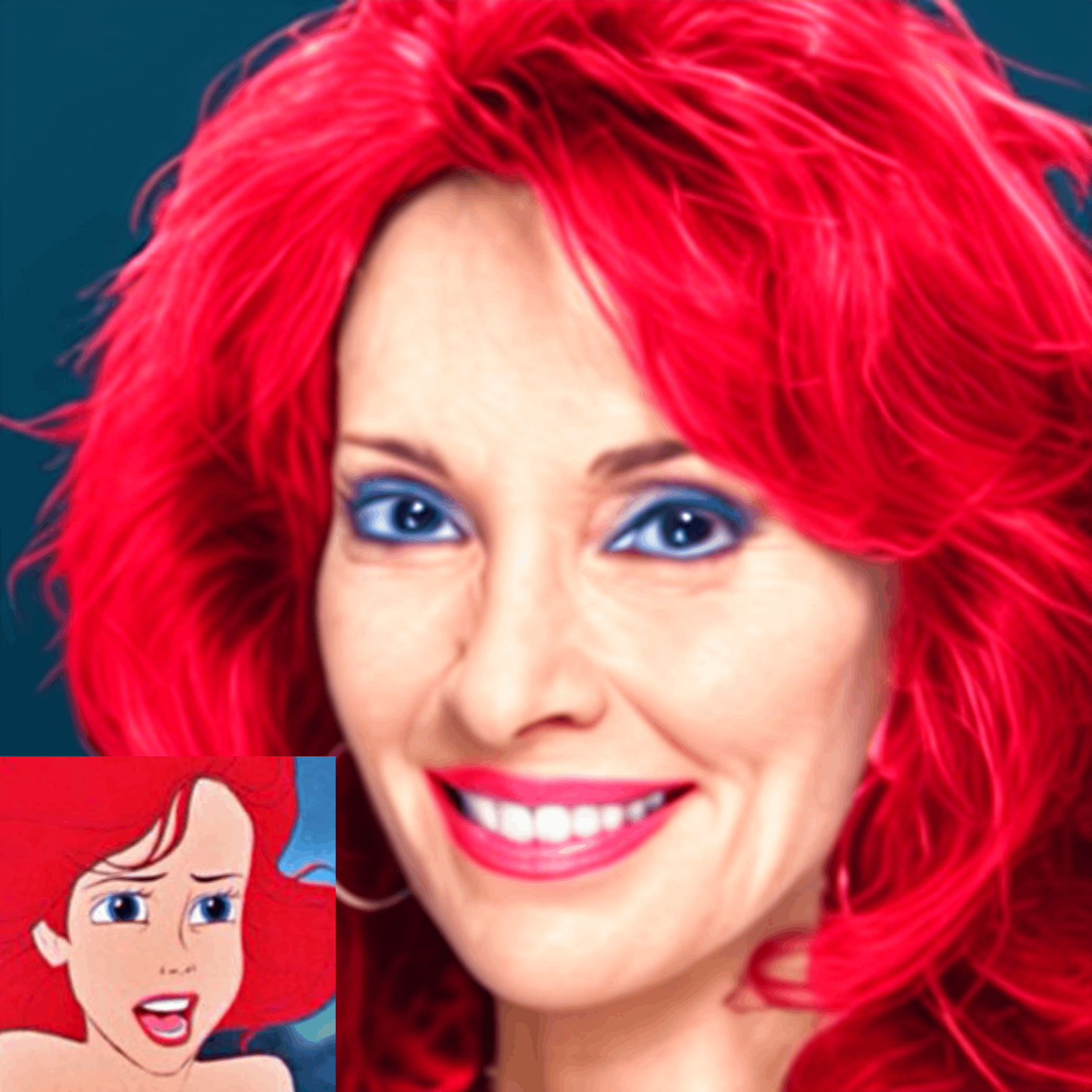}\vspace{1pt}
        \includegraphics[width=1.0\textwidth]{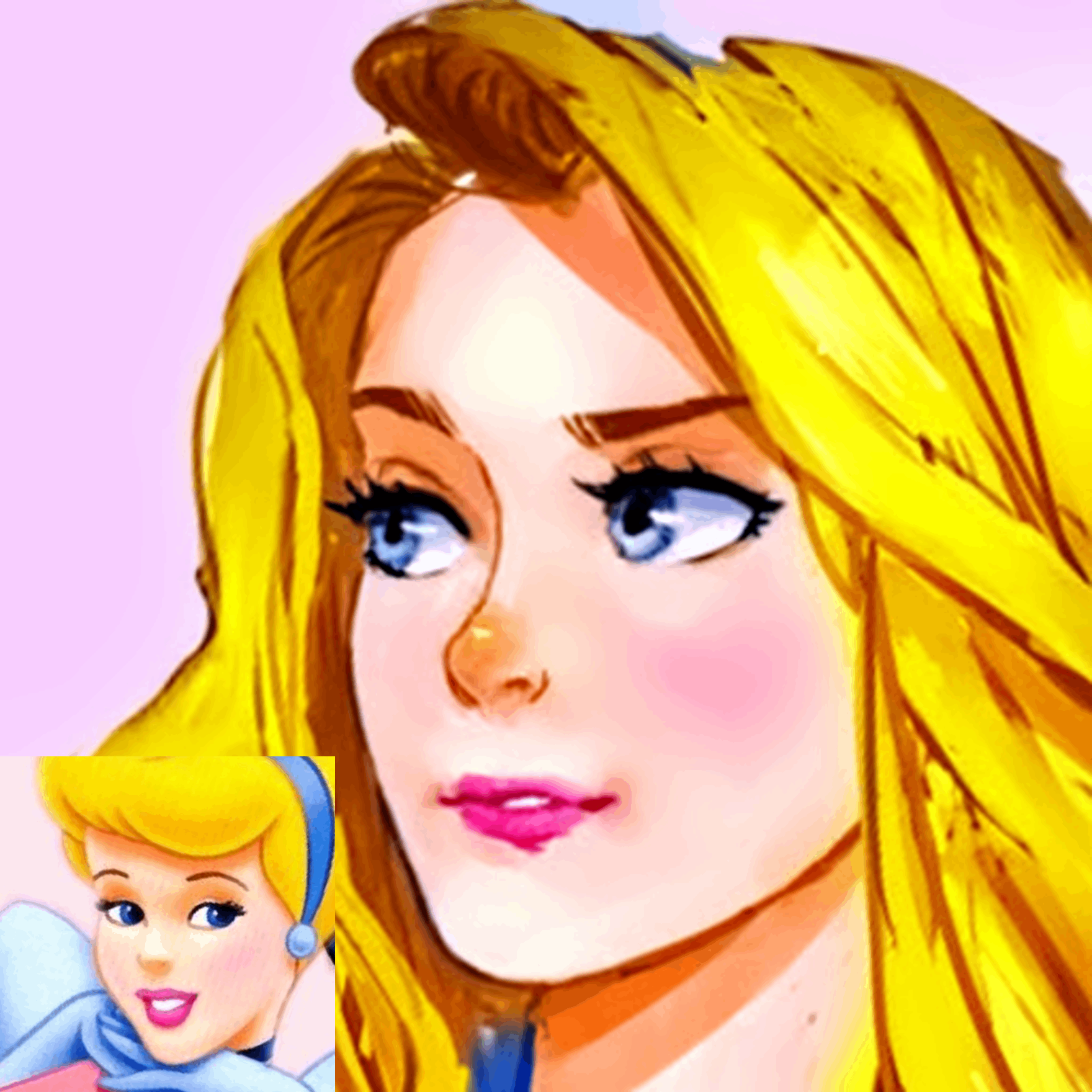}
    \end{subfigure}
    \begin{subfigure}[c]{0.160\textwidth}
        \includegraphics[width=1.0\textwidth]{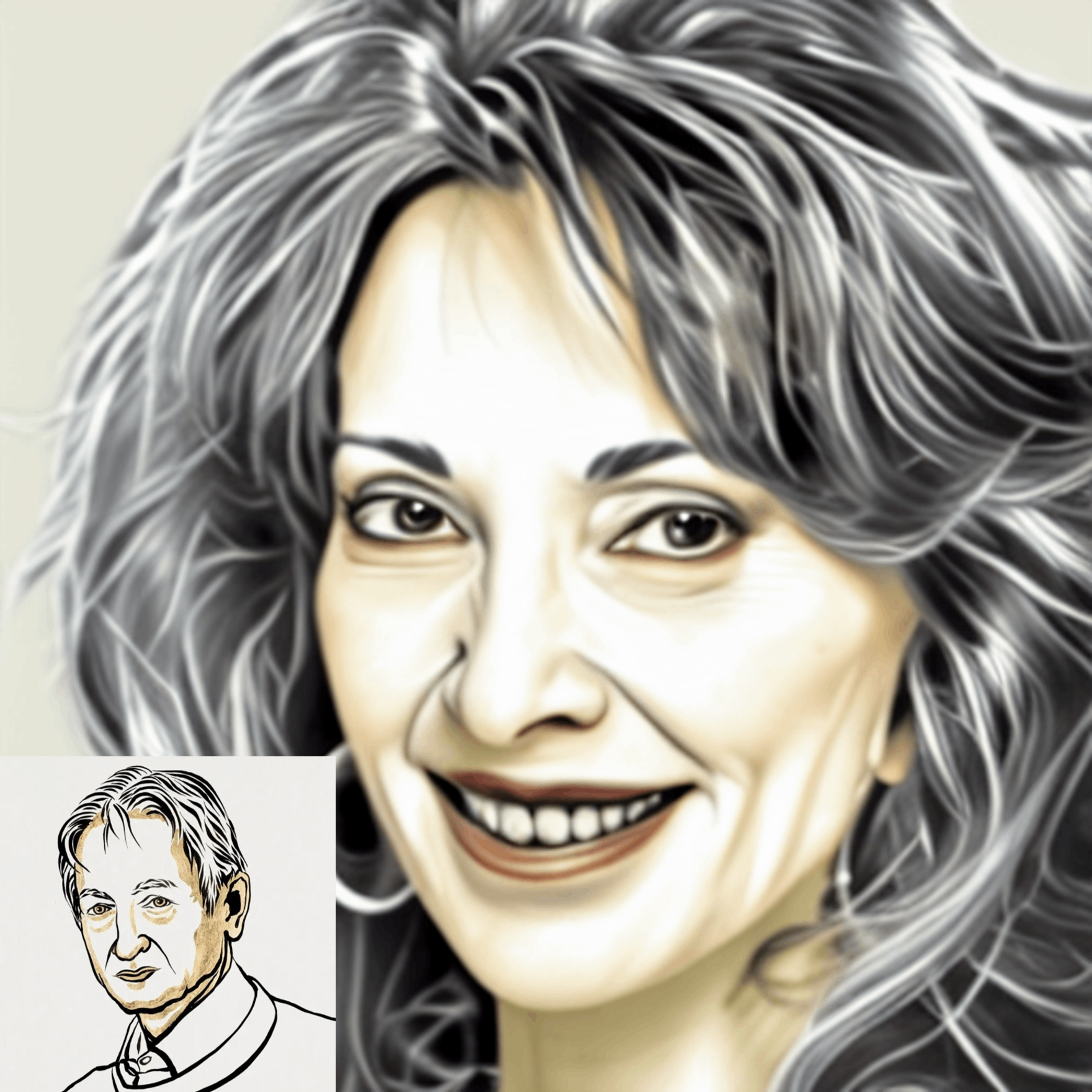}\vspace{1pt}
        \includegraphics[width=1.0\textwidth]{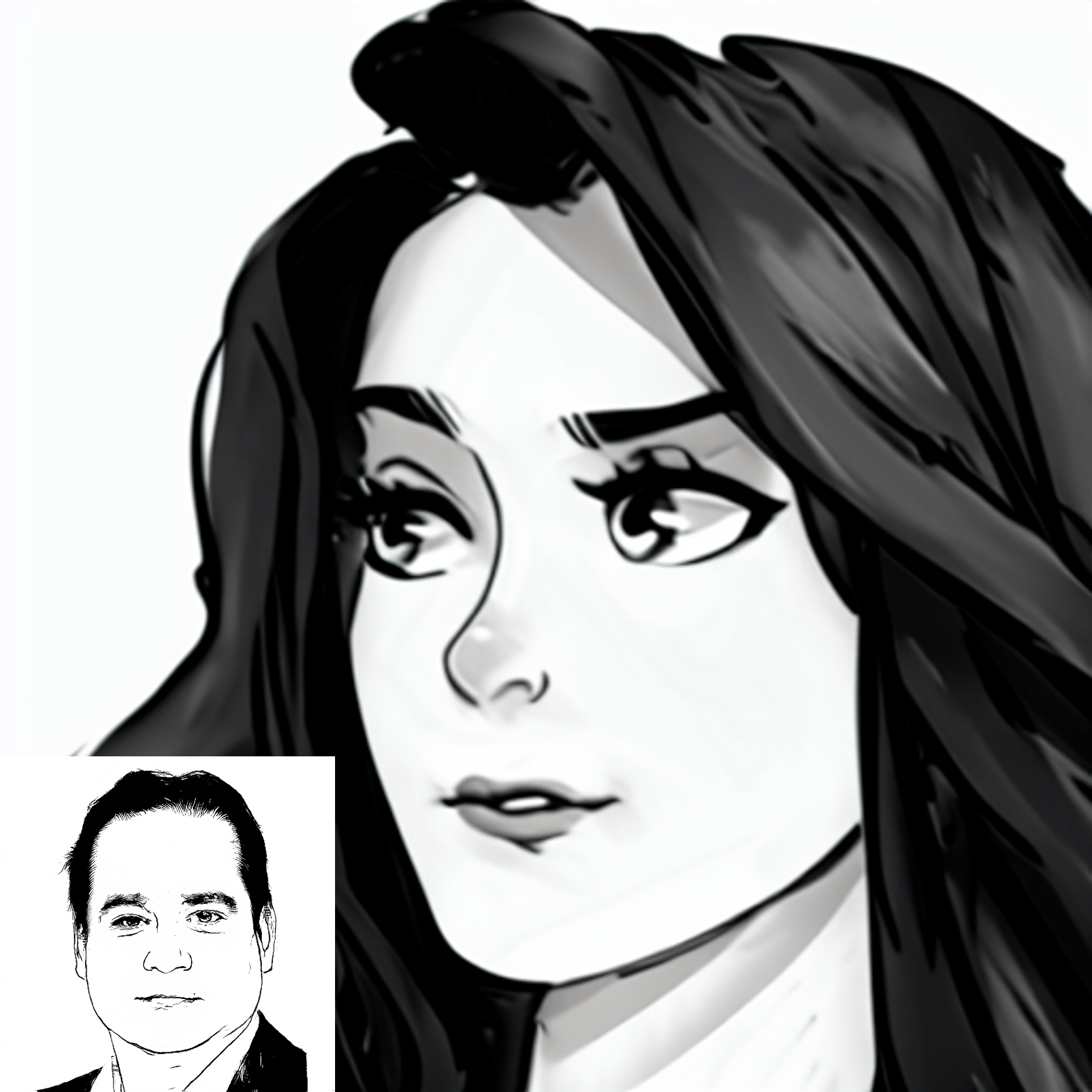}
    \end{subfigure}
    \hspace{0.025cm} 
    \begin{subfigure}[c]{0.160\textwidth}
        \includegraphics[width=1.0\textwidth]{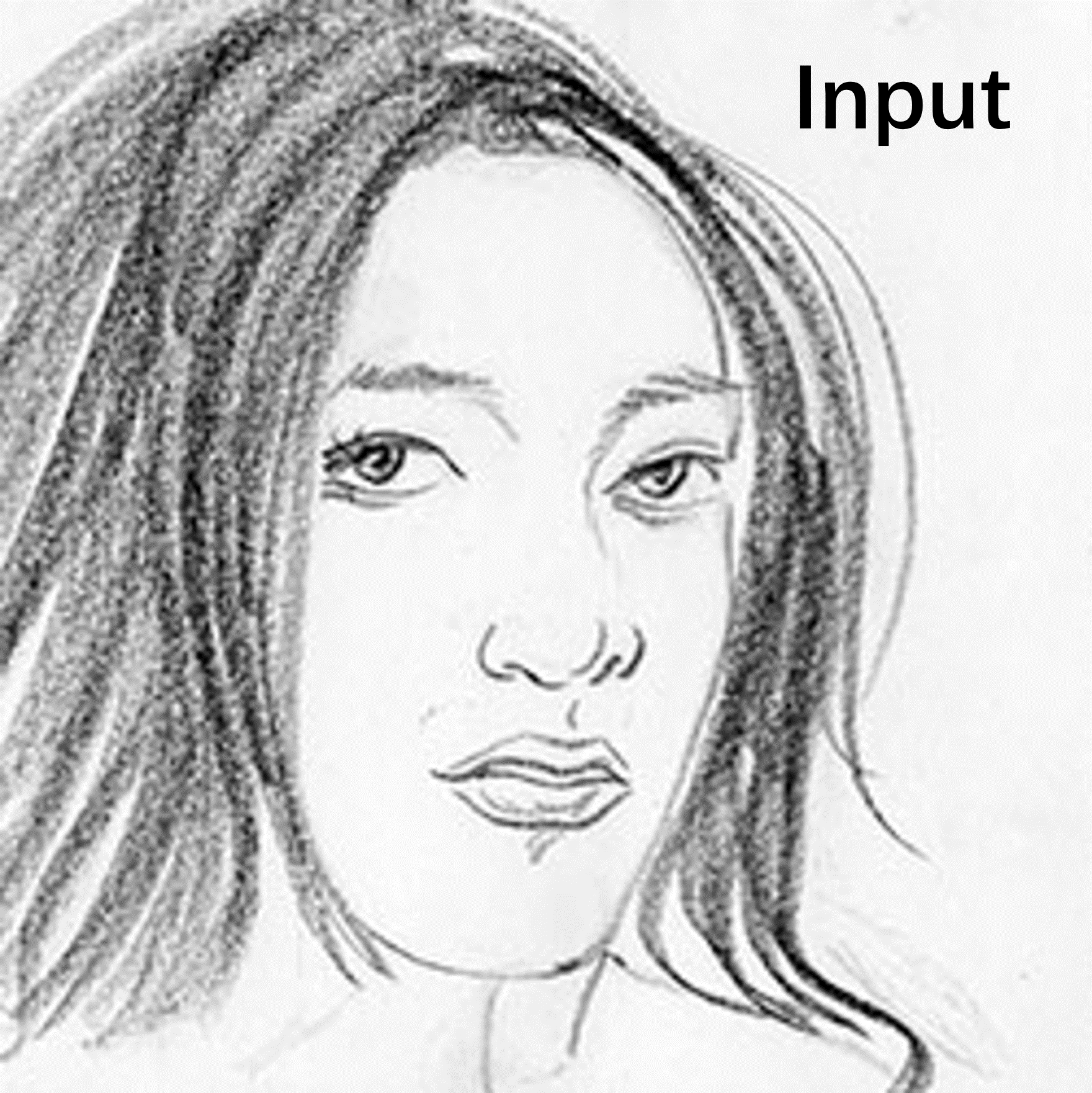}\vspace{1pt}
        \includegraphics[width=1.0\textwidth]{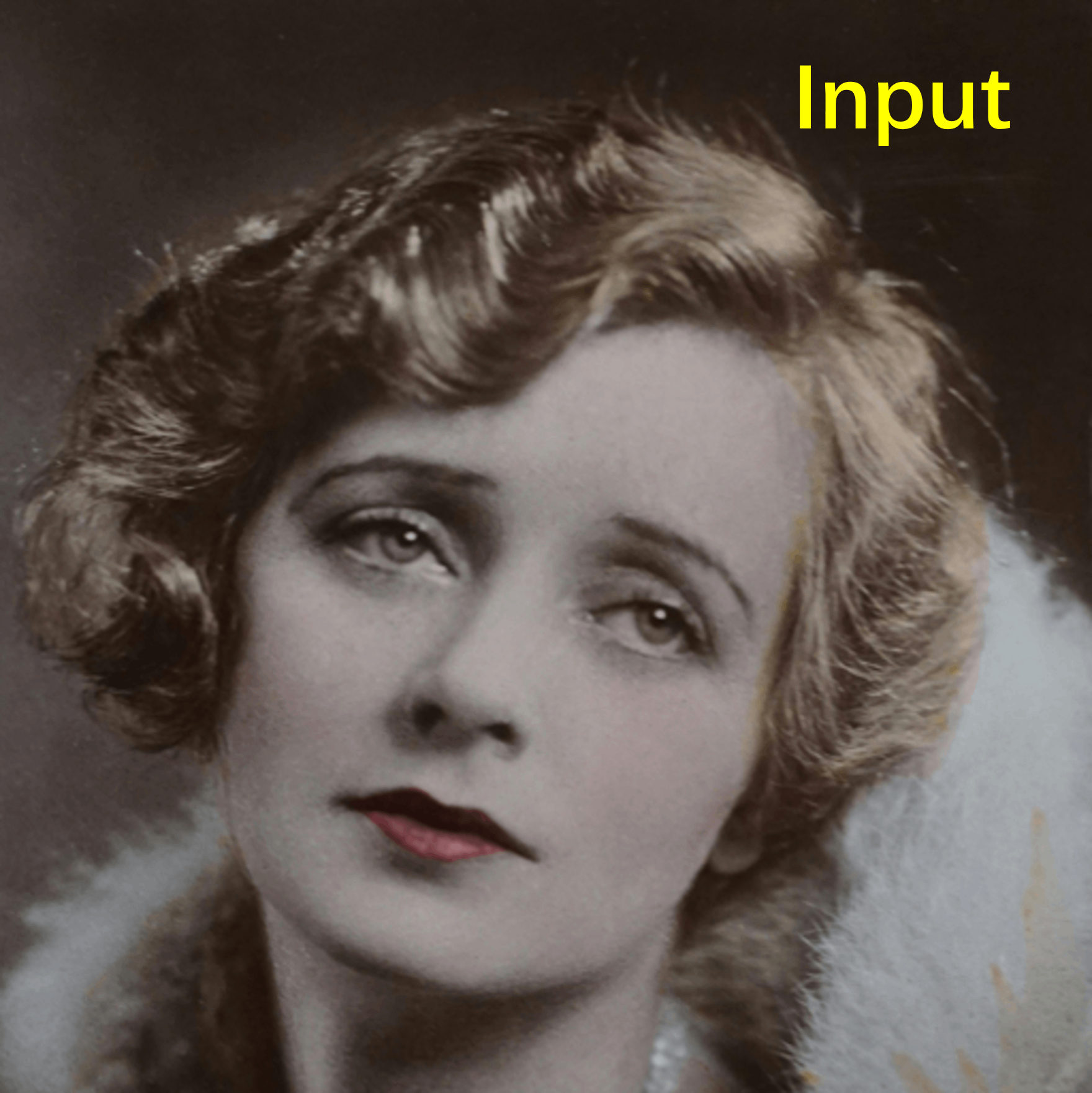}
    \end{subfigure}
    \begin{subfigure}[c]{0.160\textwidth}
    \includegraphics[width=1.0\textwidth]{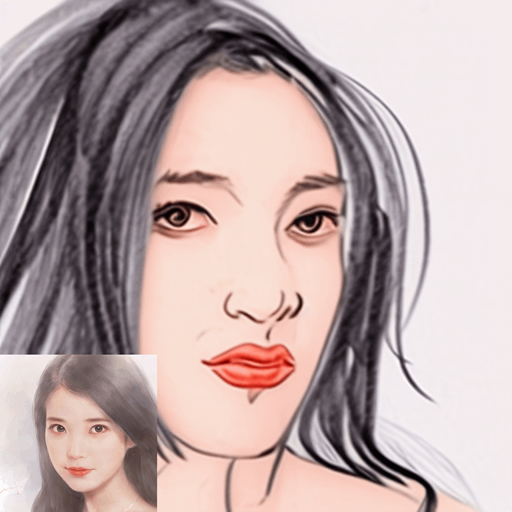}\vspace{1pt}
    
        \includegraphics[width=1.0\textwidth]{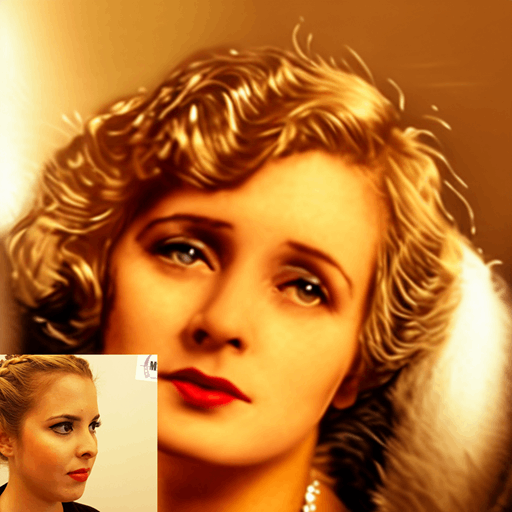}
    \end{subfigure}
    \end{minipage}
    \caption{\textbf{Domain generalizable portrait style transfer.} By training on the CelebAMask-HQ dataset with only 30K portrait photos, our method allows for high-quality semantic-aware style transfer between any two portraits from a wide variety of domains including photo, cartoon, sketch,  animation, etc. }
    \label{fig:teaser}
    \end{center}
}]
\addtocounter{footnote}{-1}

\renewcommand{\thefootnote}{\fnsymbol{footnote}}
\footnote{
\begin{tabular}{@{} p{\linewidth} @{}}
  \textsuperscript{†}Equal contribution.\\ \textsuperscript{*}Corresponding author.
\end{tabular}
} 

\begin{abstract}

This paper presents a portrait style transfer method that generalizes well to various different domains while enabling high-quality semantic-aligned stylization on regions including hair, eyes, eyelashes, skins, lips, and background. To this end, we propose to establish dense semantic correspondence between the given input and reference portraits based on a pre-trained model and a semantic adapter, with which we obtain a warped reference semantically aligned with the input. To ensure effective yet controllable style transfer, we devise an AdaIN-Wavelet transform to balance content preservation and stylization by blending low-frequency information of the warped reference with high-frequency information of the input in the latent space. A style adapter is also designed to provide style guidance from the warped reference. With the stylized latent from AdaIN-Wavelet transform, we employ a dual-conditional diffusion model that integrates a ControlNet recording high-frequency information and the style guidance to generate the final result. Extensive experiments demonstrate the superiority of our method. Our code and trained model are available at \href{https://github.com/wangxb29/DGPST}{https://github.com/wangxb29/DGPST}.

\end{abstract}    

\section{Introduction}
\label{sec:intro}

Portrait style transfer has gained popularity as an advanced image editing technique, providing an automated way to apply the visual style of a reference portrait to an input portrait. This task is particularly challenging as it requires precise local tone adjustments across different facial regions, such as the skin, lips, eyes, hair, and background, to faithfully replicate the reference style while preserving the portrait identity and facial structure details. As illustrated in Figure \ref{fig:teaser}, this technique is expected to enable a wide range of transformations, such as converting a human portrait into various artistic styles, modernizing old photographs, and adding color to sketch images, allowing casual photographers to effortlessly transform their photos into stunning masterpieces inspired by their favorite artworks.

To enhance the efficiency of portrait style transfer, several prior works \cite{shih_sig14,Gu_2019_CVPR,sun2022ide,chen2021deepfaceediting,wang2023towards,shuportraitlighting,yang2022Pastiche,zhou2024deformable,wang2025ppst} have demonstrated compelling results. However, these methods are not easily adaptable for generating portraits across multiple domains. More recently, diffusion models have been leveraged for various applications, including semantic matching \cite{tang2023emergent,zhang2024tale,li2024sd4match} and style transfer \cite{chung2023style,zhang2023inversion,wang2023stylediffusion,wang2024instantstyle,wang2024instantstyleplus,jeong2024training,deng2023zzeroshotstyletransfer,cho2024one}. While techniques such as those proposed in \cite{shih_sig14,shuportraitlighting,chen2021deepfaceediting} aim to achieve stylization by aligning semantic regions, they are effective only when the structural differences between the input and reference portraits are minimal. When applied to portraits with significant structural variations, these methods struggle to transfer style effectively between corresponding semantic regions.

In this paper, we propose a novel domain-generalizable diffusion model for portrait style transfer. Our approach first warps the reference portrait using correspondences learned from diffusion models, producing a high-quality warped style reference. To generate realistic and visually coherent portrait, our framework is designed to integrate a dual-conditional diffusion model, leveraging ControlNet to extract high-frequency details from the input image for structural guidance, while an image adapter utilizes the warped reference to provide style guidance.

Unlike previous diffusion-based methods \cite{zhang2023inversion,wang2024instantstyleplus,chung2023style,deng2023zzeroshotstyletransfer,cho2024one}, our approach establishes dense semantic-aligned correspondences to enhance the quality of style transfer. Specifically, we introduce a semantic adapter, comprising a frozen diffusion U-Net and a CLIP encoder to extract image features from both the content and style reference images. As naively using these extracted features for correspondence estimation can result in incomplete semantic regions due to misalignment, we introduce mask warping loss and cyclic warping consistency loss for optimizing the semantic adapter to encourage semantic consistency in the extracted features. With this refined representation, our method enables accurate semantic-aware style transfer while ensuring the visual coherence in the generated portraits.


Color tone plays a crucial role in defining artistic style during portrait style transfer. If the generation process starts from the latent representation of the input image, the resulting portrait tends to retain its original color tone, limiting the effectiveness of style transfer. To better integrate structure and style, we propose a novel AdaIN-Wavelet Transform for latent initialization. 
Specifically, we first compute the latent representation of the warped reference image using DDIM inversion, as initializing from this latent naturally enhances color transfer effect. However, since the warped reference image may lose some fine-scale content details, directly using its latent representation may result in blurred portrait. To mitigate this, we blend the low-frequency information of the warped reference latent with the high-frequency information of the input latent, achieving a balance between stylization and content preservation. 

With the above mentioned designs, our model enables effective portrait style transfer across various different domains. Through extensive experiments, we demonstrate the effectiveness of our approach in various artistic styles, modernizing old photographs, and colorizing sketch images. Our key contributions can be summarized as follows:


\begin{itemize}
 \setlength\itemsep{0.5em}
    \item We present a portrait style transfer framework that can robustly process portraits from various different domains by training on a small-scale real-world portrait dataset.  
    \item We propose to utilize pre-trained diffusion model and also a semantic adapter for robust dense correspondence extraction. In addition, we develop a AdaIN-Wavelet transform to balance stylization and content preservation.
    \item Extensive experiments show that our method significantly outperforms previous methods in both visual quality and quantitative metrics.
\end{itemize}

\begin{figure*}[t]
    \vspace{-2mm}
	\hsize=\textwidth
	\centering
	\includegraphics[width=1.0\textwidth]{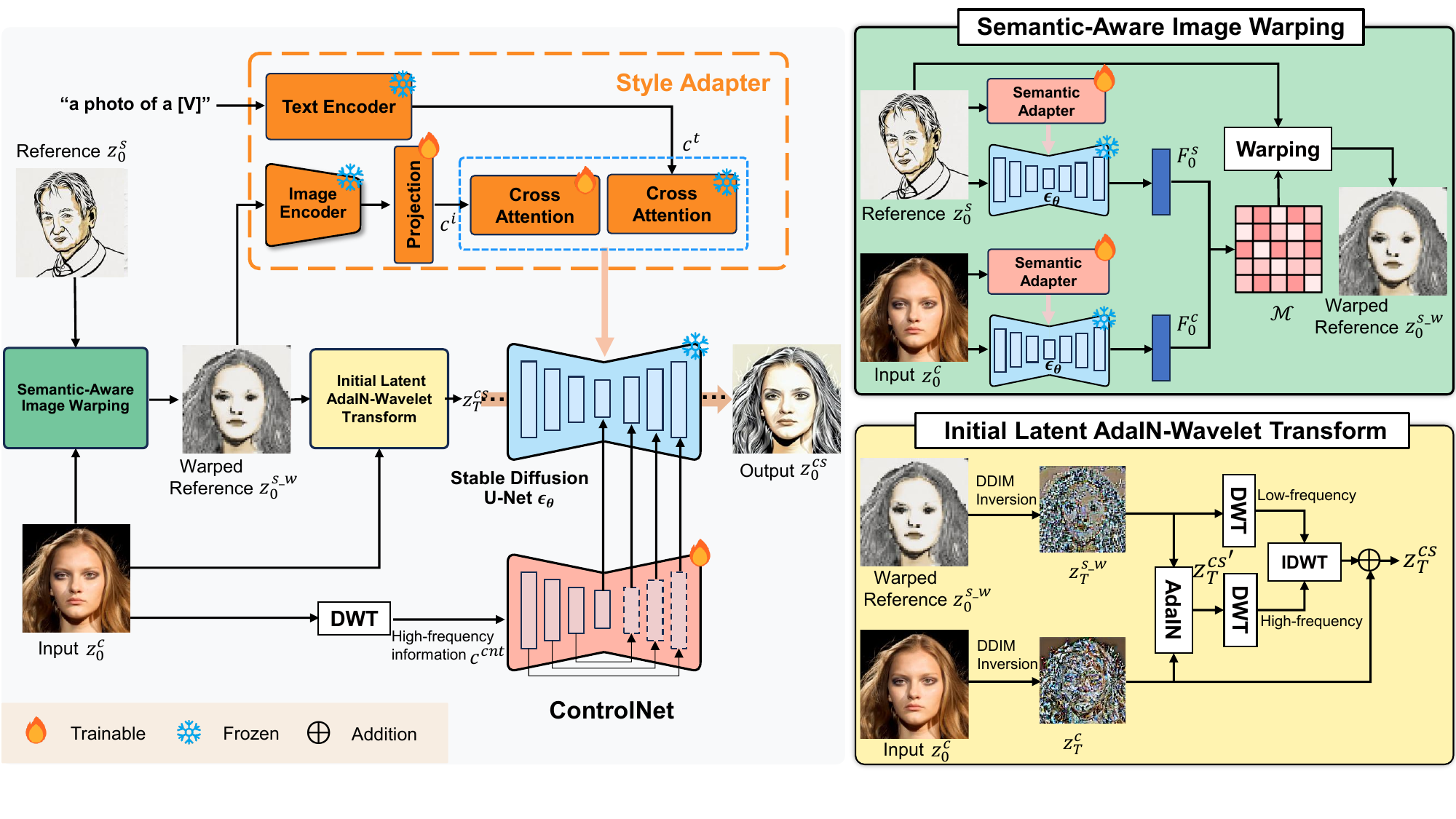} \\
     \vspace{-4mm}
\caption{\textbf{Framework overview.} (Left) Illustration of the proposed portrait style transfer method. We first feed the input and reference portraits into Stable Diffusion and the Semantic Adapter. Next, we extract diffusion features and compute a correlation matrix, which is used to warp the reference portrait. During sampling, we provide structure guidance by extracting the high-frequency information of the input portrait using Discrete Wavelet Transform and feed it into ControlNet. For style guidance, the warped reference is input into the style adapter. To obtain the initial latent, we perform DDIM inversion on both the warped reference and input portraits, followed by applying AdaIN between them. The final initial latent is obtained by combining the low-frequency information of the warped reference latent with the high-frequency information from the input latent after AdaIN.} 
\label{pipiline}
\vspace{-2mm}
\end{figure*}

\section{Related Work}
\label{sec:Related_Work}
\noindent \textbf{Diffusion-based Style Transfer.} Diffusion models have been explored for style transfer, with a common approach leveraging cross-attention layers in text-to-image diffusion models. These methods can be broadly categorized into three groups: (i) directly using text prompts \cite{wang2024instantid}, (ii) mapping the reference image into textual embeddings \cite{zhang2023inversion,zhang2023prospect}, and (iii) incorporating an additional cross-attention mechanism with image prompts \cite{ye2023ip,wang2024instantstyle,qi2024deadiff}. Beyond these approaches, training-free methods have also been developed based on pre-trained text-to-image diffusion models. These methods inject features into self-attention layers \cite{chung2023style,deng2023zzeroshotstyletransfer,lin2024ctrl} during the generation process to guide stylization. Additionally, StyleDiffusion \cite{wang2023stylediffusion} fine-tunes a diffusion model for a given reference image using a CLIP-based style disentanglement loss. However, most existing diffusion-based methods are designed for artistic style transfer that does not need to consider semantic correspondence, and thus basically struggle to deal with portrait style transfer involving strong semantic relevance, especially when the portraits are from different domains. 


\vspace{0.5em}
\noindent \textbf{Portrait Style Transfer.} Significant progress has recently been made in this field. \citet{shih_sig14} pioneered photorealistic portrait style transfer by first warping the reference portrait to align with the input portrait, followed by a multiscale local contrast transfer. Concurrently, \citet{shuportraitlighting} redefined portrait relighting as a mass transport problem and proposed an algorithm to address it. With the advent of CNN-based neural style transfer \cite{Gatys_2016_CVPR,luan2017deep}, specialized methods for portrait style transfer emerged, including photorealistic \cite{guo2019robust} and artistic styles \cite{selim2016painting}. Among them, \citet{guo2019robust} enhanced spatial transformations by maximizing the cross-correlation of features within segmented regions between the input and reference images. GAN-based methods have also gained popularity in face stylization due to their ability to generate high-quality, highly detailed images \cite{Gu_2019_CVPR,chen2021deepfaceediting,wang2023towards}. Notably, some approaches leverage semantic masks to enable style transfer for specific facial regions \cite{wang2023towards,Gu_2019_CVPR}. Moreover, StyleGAN \cite{StyleGAN,Karras2019stylegan2}, known for its impressive high-resolution face synthesis and hierarchical style control, has served as a foundation for several recent artistic portrait style transfer methods \cite{yang2022Pastiche,Song2021AgileGAN,VToonify,zhou2024deformable}. However, these methods are unsuitable for our task, as they inevitably alter portrait identity.

\section{Preliminary}
\noindent \textbf{Diffusion model} is a class of generative models consisting of a $T$-step forward process and a $T$-step
reverse process. Here, $T$ represents the total number of steps in the forward process. At each timestep $t$ , the forward process adds Gaussian noise to the input according to a predetermined noise schedule $\alpha_t$:
\begin{equation}
	z_t = \sqrt{\alpha_t}\cdot z_0 + \sqrt{1-\alpha_t}\cdot \epsilon, \epsilon \sim \mathcal{N}(0,I).
\end{equation}
A corresponding reverse process involves predicting the added noise $\epsilon$ using a network $\epsilon_{\theta}$, which takes $z_t$ and timestep $t$ as input, and then samples $z_{t-1}$ from $z_t$ using a formula, such as DDPM \cite{ho2020denoising} and DDIM \cite{song2020denoising}. The $\epsilon_{\theta}$ is usually parameterized as a U-Net for image generation. The added noise is progressively removed, ultimately resulting in a clean image. In this paper, we implement our method using Stable Diffusion (SD) \cite{rombach2022high}, a large-scale pre-trained text-to-image model. The training objective of SD is defined as:
\begin{equation}
\mathcal{L} = \mathbb{E}_{z_t, t, y, \epsilon \sim \mathcal{N}(0,1)} \left[ \| \epsilon - \epsilon_\theta (z_t, t, y) \|_2^2 \right],
\end{equation}
where $y$ is the embedding of text prompts generated by a pre-trained CLIP \cite{radford2021learning} text encoder.

\begin{figure*}[t]
	\captionsetup[subfloat]{labelsep=none,format=plain,labelformat=empty}
    \begin{subfigure}[c]{0.12\textwidth}
        \centering
        \includegraphics[width=1.0\textwidth]{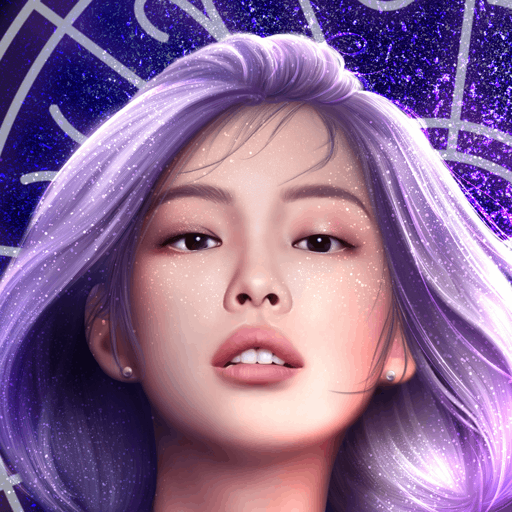}\vspace{1pt}
        \includegraphics[width=1.0\textwidth]{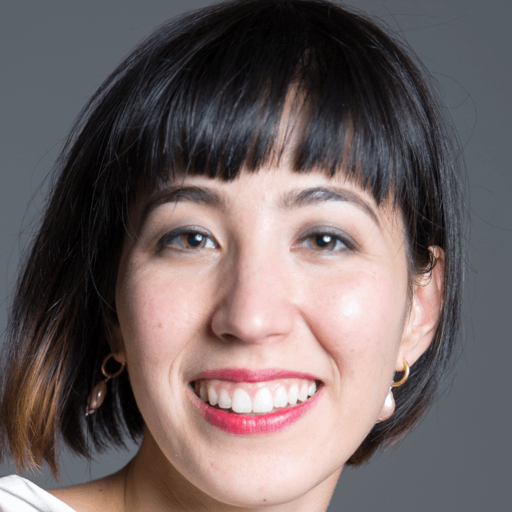}\vspace{1pt}
        \includegraphics[width=1.0\textwidth]{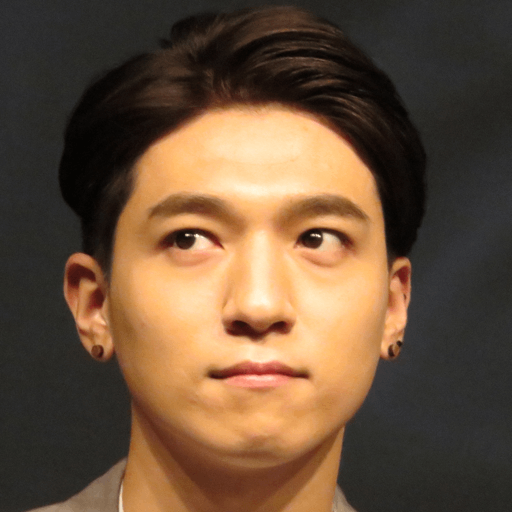}\vspace{1pt}
        \includegraphics[width=1.0\textwidth]{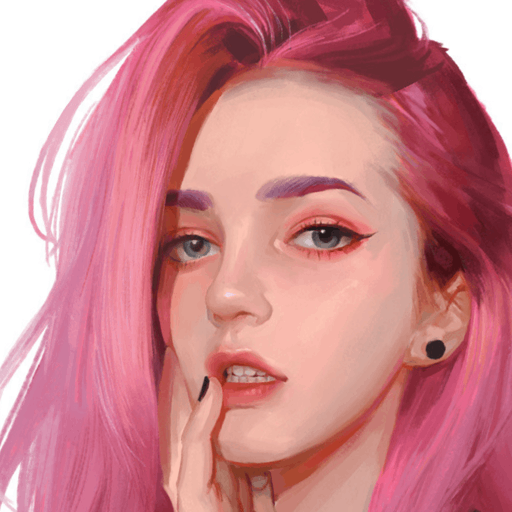}\vspace{1pt}
        \includegraphics[width=1.0\textwidth]{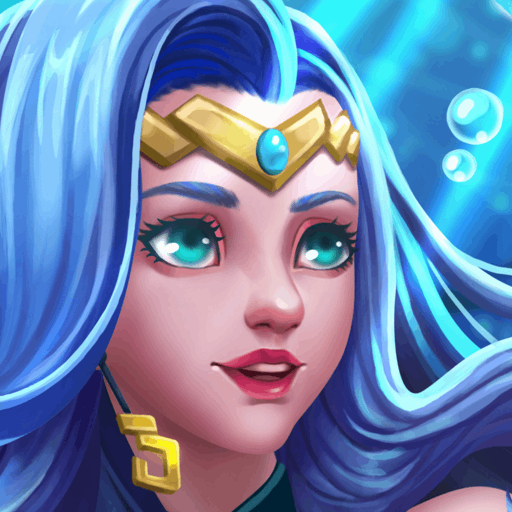}\vspace{1pt}
        \caption{Reference}
    \end{subfigure}
    \begin{subfigure}[c]{0.12\textwidth}
        \centering
        \includegraphics[width=1.0\textwidth]{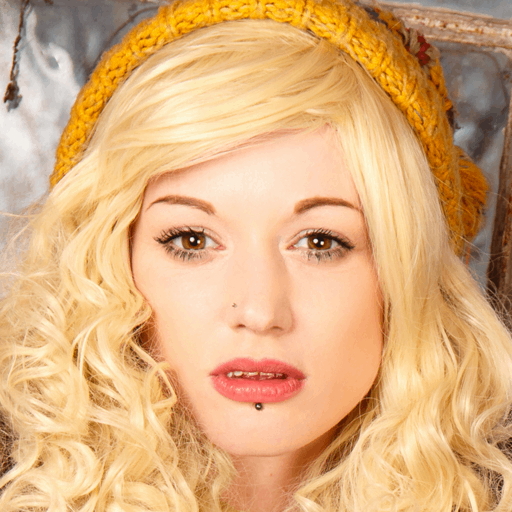}\vspace{1pt}
        \includegraphics[width=1.0\textwidth]{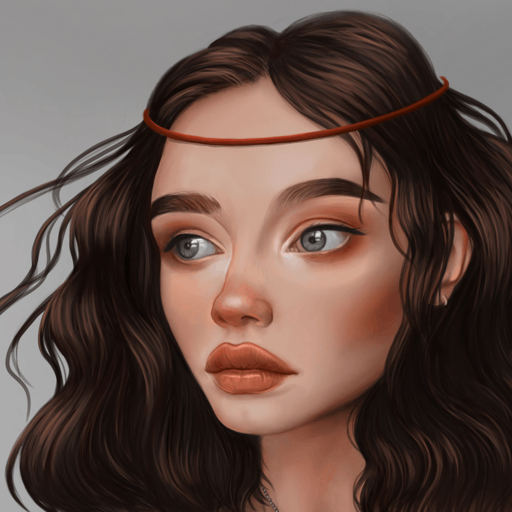}\vspace{1pt}
        \includegraphics[width=1.0\textwidth]{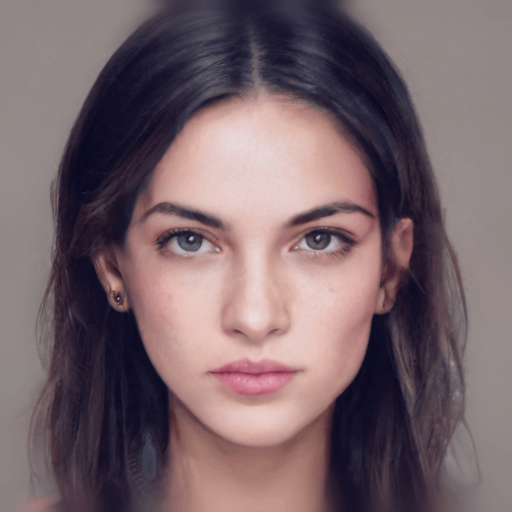}\vspace{1pt}
        \includegraphics[width=1.0\textwidth]{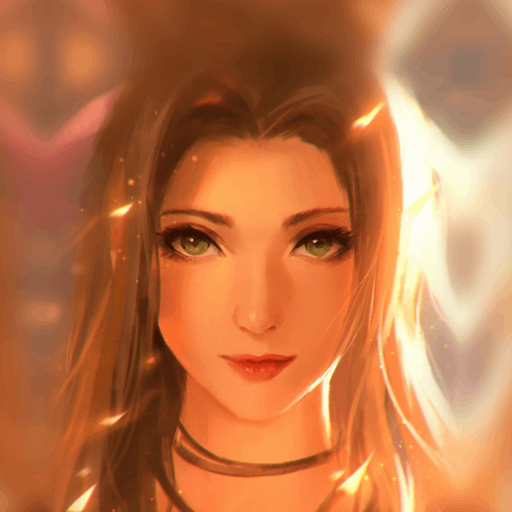}\vspace{1pt}
        \includegraphics[width=1.0\textwidth]{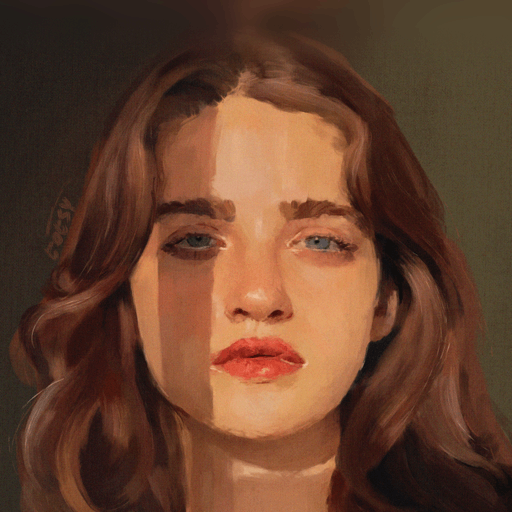}\vspace{1pt}
        \caption{Input}
    \end{subfigure}
    \begin{subfigure}[c]{0.12\textwidth}
        \centering
        \includegraphics[width=1.0\textwidth]{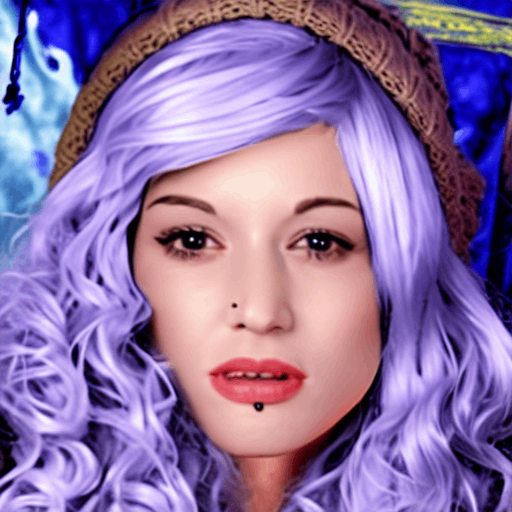}\vspace{1pt}
        \includegraphics[width=1.0\textwidth]{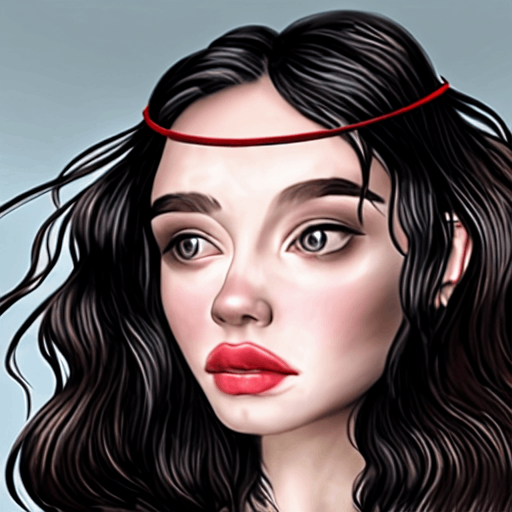}\vspace{1pt}
        \includegraphics[width=1.0\textwidth]{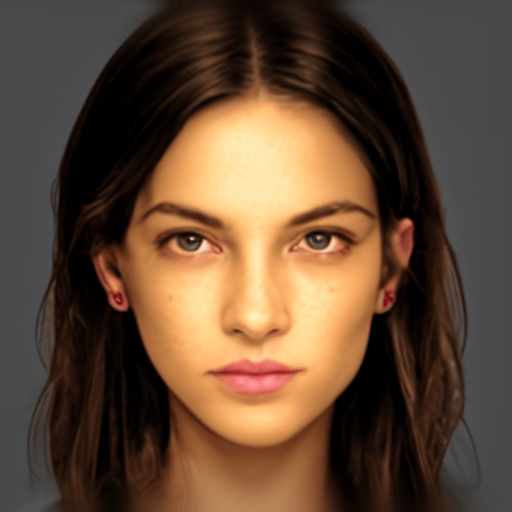}\vspace{1pt}
        \includegraphics[width=1.0\textwidth]{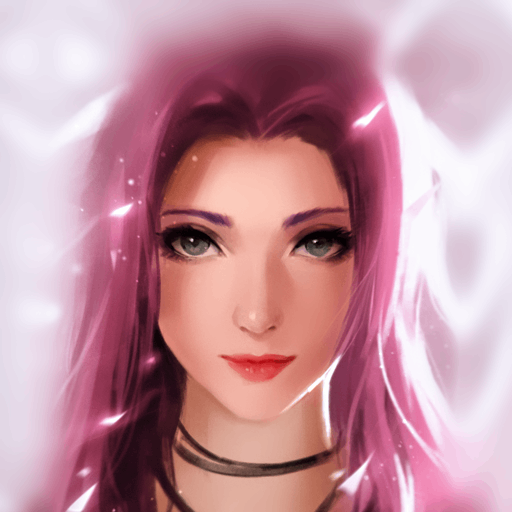}\vspace{1pt}
        \includegraphics[width=1.0\textwidth]{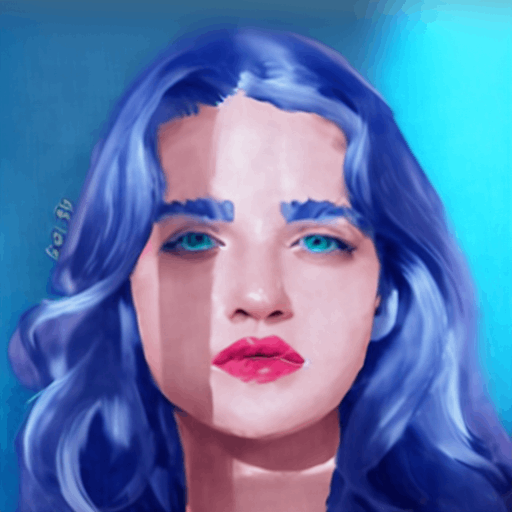}\vspace{1pt}
        \caption{Ours}
    \end{subfigure}
    \begin{subfigure}[c]{0.12\textwidth}
        \centering
        \includegraphics[width=1.0\textwidth]{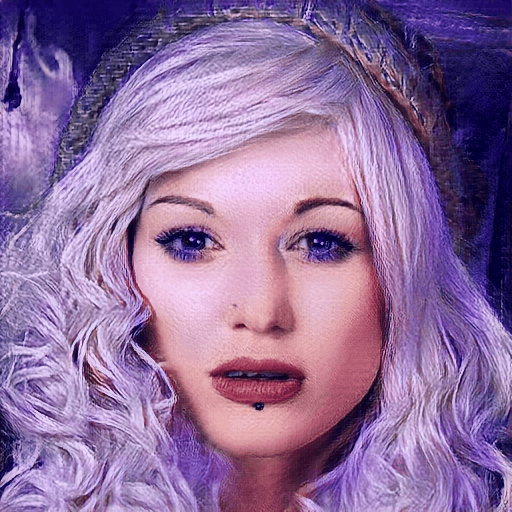}\vspace{1pt}
        \includegraphics[width=1.0\textwidth]{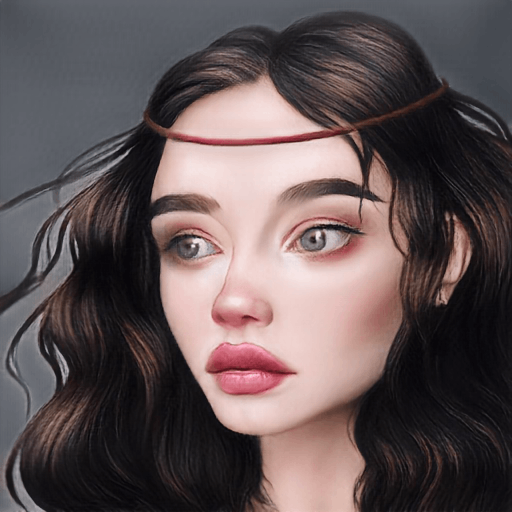}\vspace{1pt}
        \includegraphics[width=1.0\textwidth]{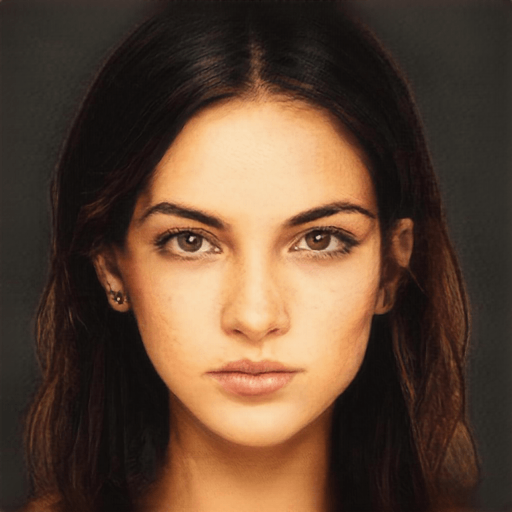}\vspace{1pt}
        \includegraphics[width=1.0\textwidth]{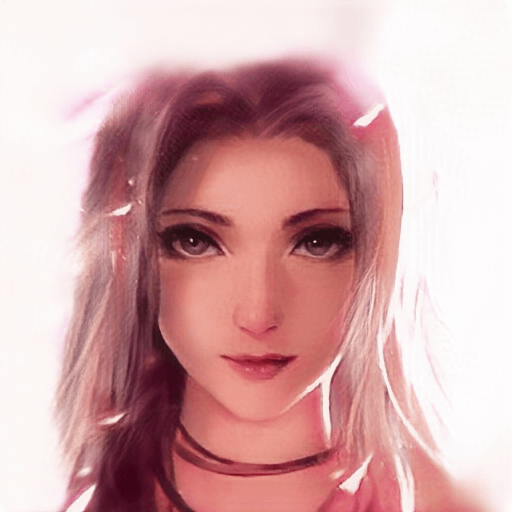}\vspace{1pt}
        \includegraphics[width=1.0\textwidth]{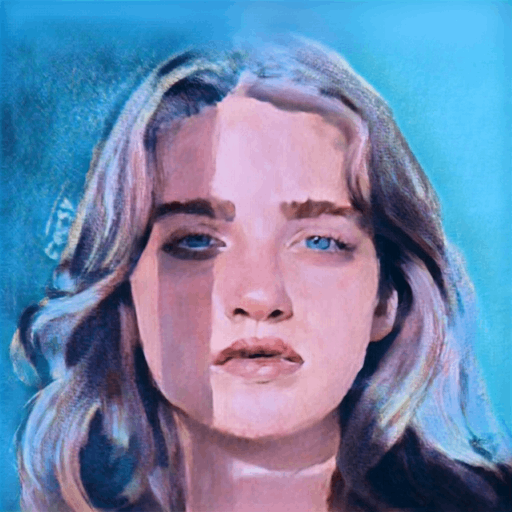}\vspace{1pt}
        \caption{\scriptsize Wang et al.}
    \end{subfigure}
    \begin{subfigure}[c]{0.12\textwidth}
        \includegraphics[width=1.0\textwidth]{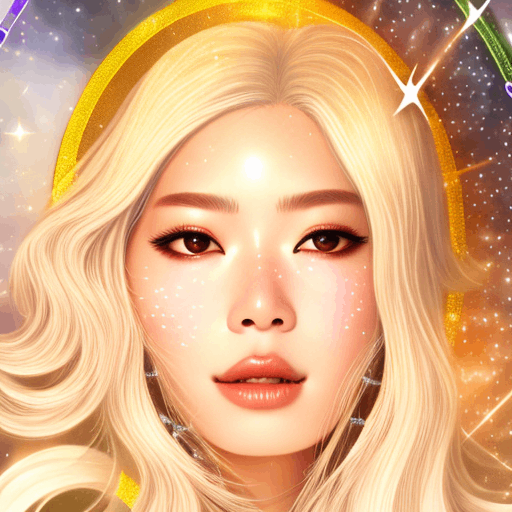}\vspace{1pt}
        \includegraphics[width=1.0\textwidth]{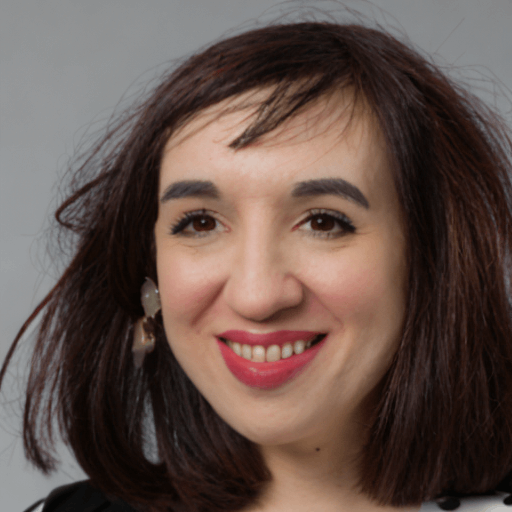}\vspace{1pt}
        \includegraphics[width=1.0\textwidth]{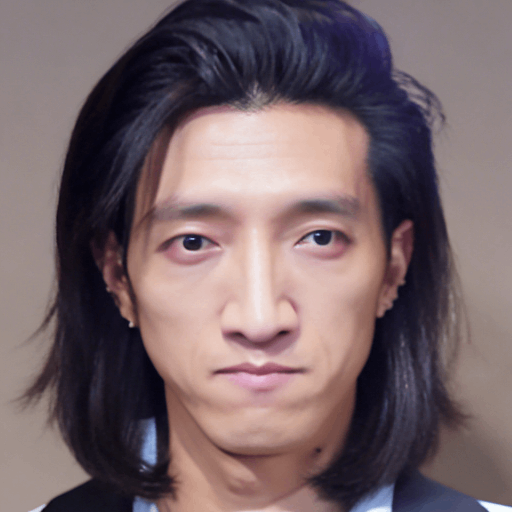}\vspace{1pt}
        \includegraphics[width=1.0\textwidth]{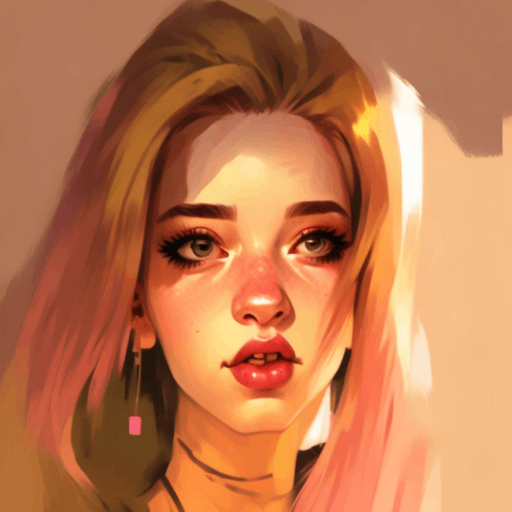}\vspace{1pt}
        \includegraphics[width=1.0\textwidth]{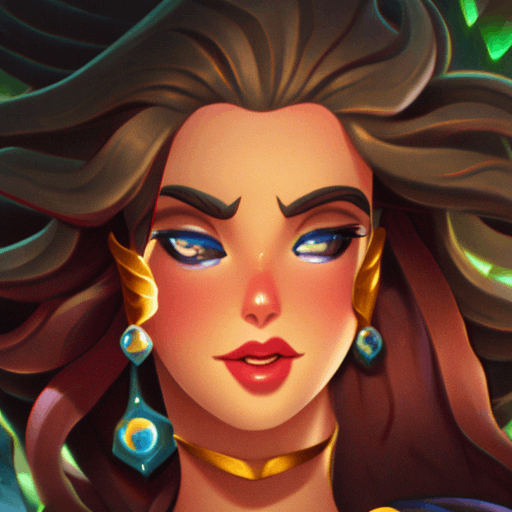}\vspace{1pt}
        \caption{\scriptsize IP-A+C.N.}
    \end{subfigure}
    \begin{subfigure}[c]{0.12\textwidth}
        \includegraphics[width=1.0\textwidth]{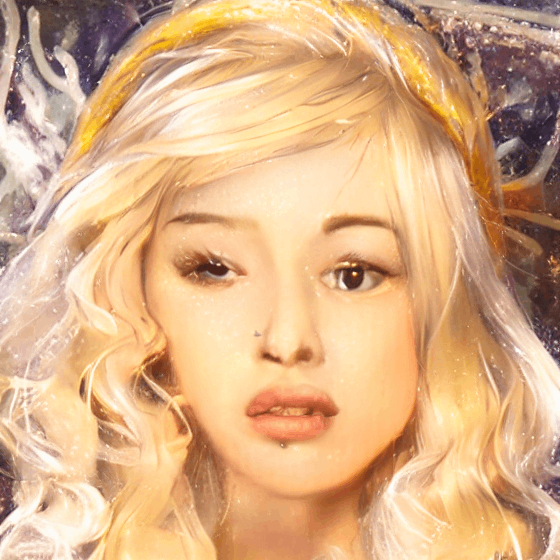}\vspace{1pt}
        \includegraphics[width=1.0\textwidth]{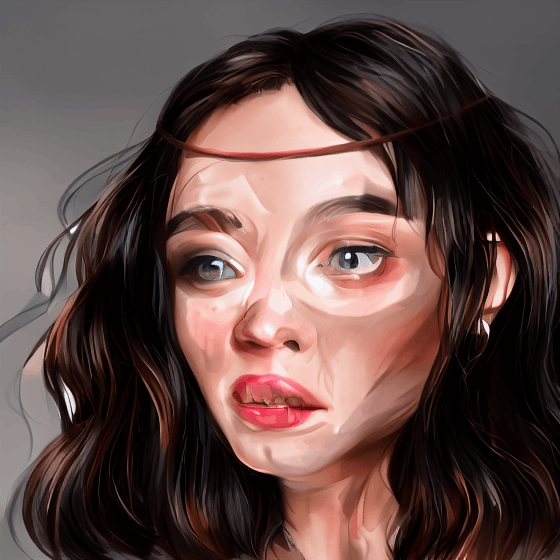}\vspace{1pt}
        \includegraphics[width=1.0\textwidth]{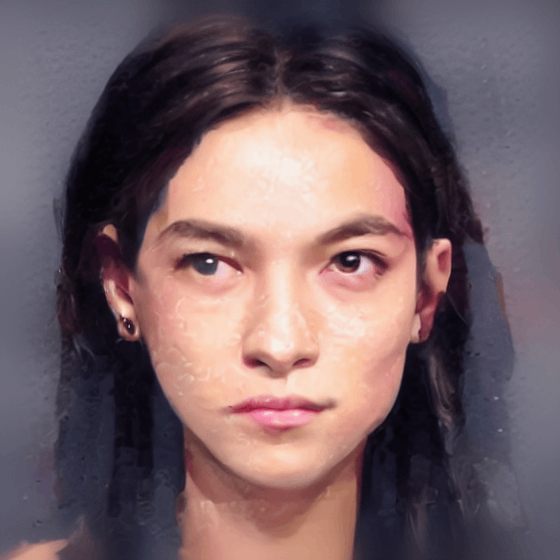}\vspace{1pt}
        \includegraphics[width=1.0\textwidth]{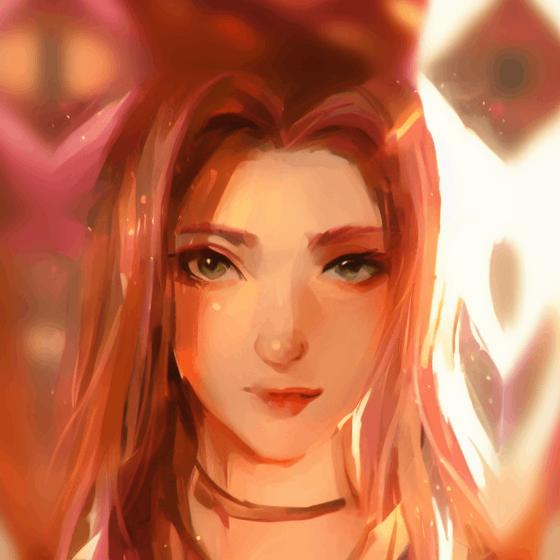}\vspace{1pt}
        \includegraphics[width=1.0\textwidth]{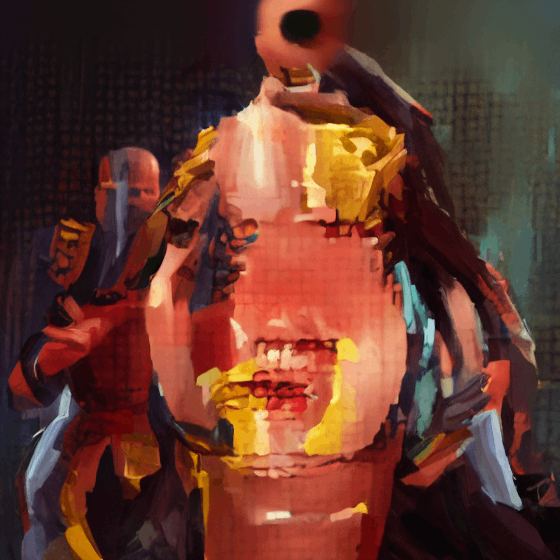}\vspace{1pt}
        \caption{\scriptsize Deng et al.}
    \end{subfigure}
    \begin{subfigure}[c]{0.12\textwidth}
        \includegraphics[width=1.0\textwidth]{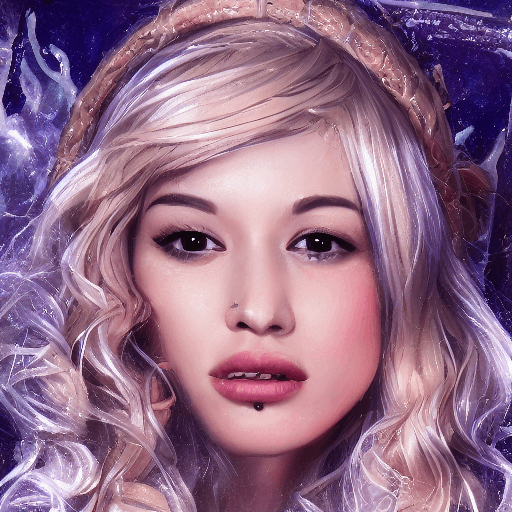}\vspace{1pt}
        \includegraphics[width=1.0\textwidth]{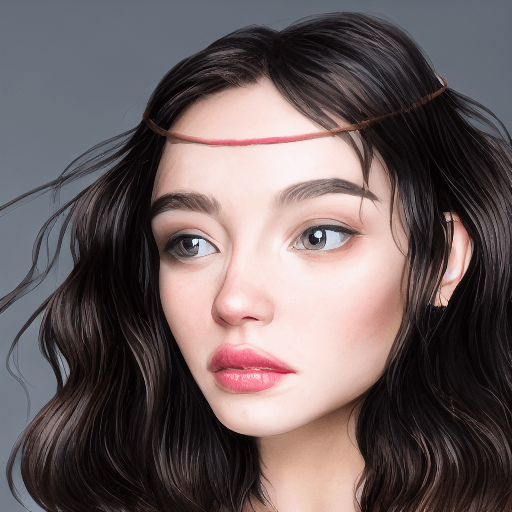}\vspace{1pt}
        \includegraphics[width=1.0\textwidth]{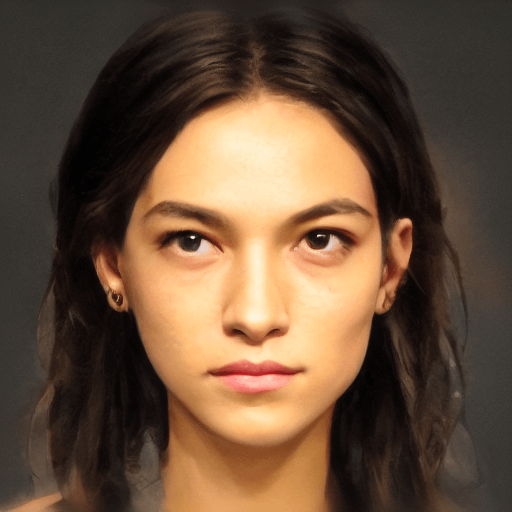}\vspace{1pt}
        \includegraphics[width=1.0\textwidth]
        {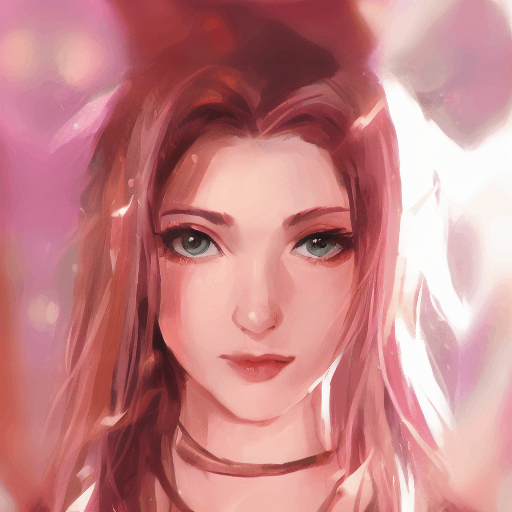}\vspace{1pt}
        \includegraphics[width=1.0\textwidth]{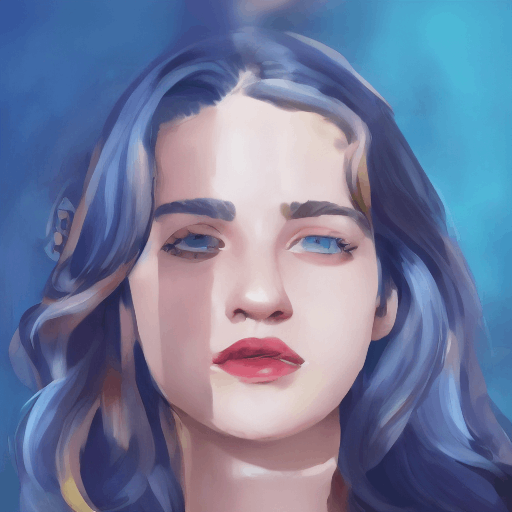}\vspace{1pt}
        \caption{\scriptsize StyleID}
    \end{subfigure}
    \begin{subfigure}[c]{0.12\textwidth}
        \includegraphics[width=1.0\textwidth]{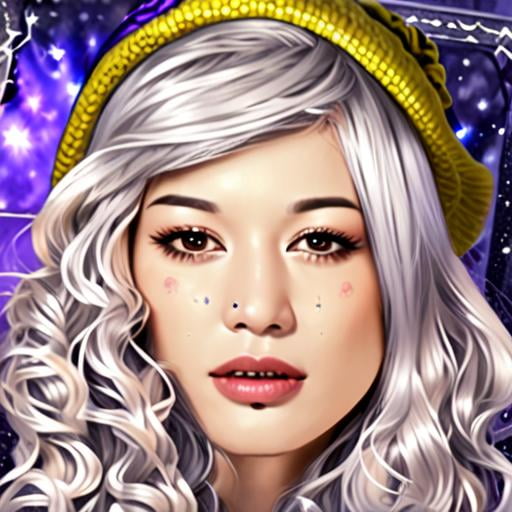}\vspace{1pt}
        \includegraphics[width=1.0\textwidth]{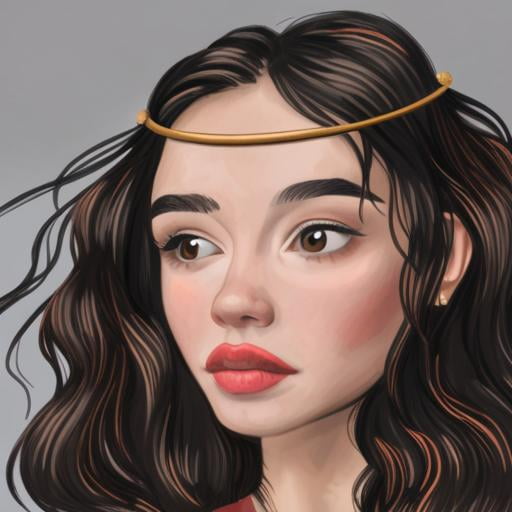}\vspace{1pt}
        \includegraphics[width=1.0\textwidth]{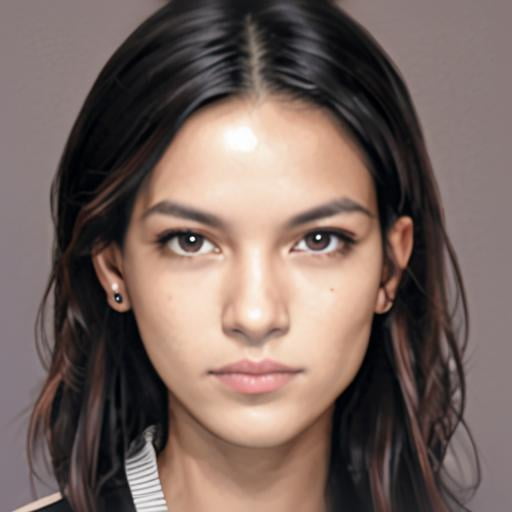}\vspace{1pt}
        \includegraphics[width=1.0\textwidth]
        {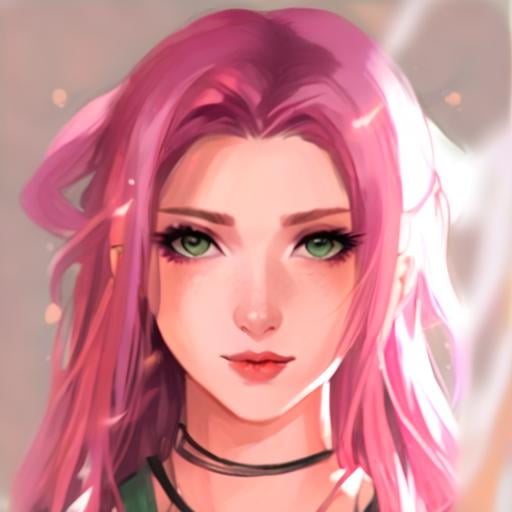}\vspace{1pt}
        \includegraphics[width=1.0\textwidth]{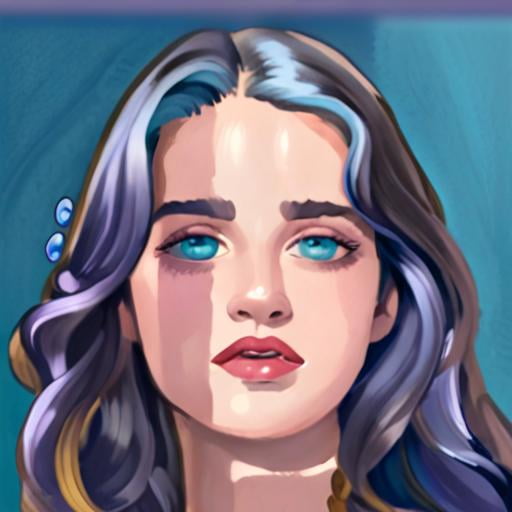}\vspace{1pt}
        \caption{\scriptsize InstantStyle+}
    \end{subfigure}
    \vspace{-2mm}
	\caption{\textbf{Qualitative comparison with SOTA portrait style transfer methods.}}
  \vspace{-2mm}
	\label{fig:comparison-conventional}
\end{figure*}	

\section{Method}
Our method aims to achieve semantically aligned portrait style transfer while preserving content details. Since our focus is on detailing the proposed method within the diffusion process, we denote the content image as $z^c_0$ and the style image as $z^s_0$. As illustrated in Figure \ref{pipiline}, our approach first warps the portrait style image using the correspondence learned from diffusion models to obtain a high-quality warped style reference $z^{s\_w}_0$ (Section \ref{corr-refin}). To generate a realistic portrait, our framework incorporates a dual-conditional diffusion model (Section \ref{ContentGuide}). This model utilizes the ControlNet $\epsilon_{\theta}$, which extracts high-frequency information from the content image $z^c_0$ to provide content guidance, while an image adapter leverages the warped reference $z^{s\_w}_0$ to offer style guidance. Additionally, we introduce a latent initialization strategy (Section \ref{Initlatent}) to achieve a better balance between stylization and content preservation.

\subsection{Semantic-Aware Style Alignment \label{corr-refin}}

Different from previous diffusion-based methods \cite{tang2023emergent, zhang2024tale} for correspondence extraction, our approach focuses on establishing dense correspondence across portraits to achieve high-quality portrait style transfer. By ensuring style transfer between semantically aligned regions, we enhance both accuracy and visual coherence. This section details our method for constructing robust semantic correspondence to enable precise and effective style alignment.

\vspace{0.5em}
\noindent \textbf{Correspondence Extraction.} 
We first utilize a semantic adapter and a pre-trained CLIP image encoder to extract image features. Specifically, we obtain features from the penultimate layer of the encoder, which consists of a total of $16\times16 +1 = 256$ tokens with one class token and 256 tokens corresponding to individual image patches. These tokens are then processed through a projection network and fed into Stable Diffusion U-Net for decoupled cross-attention \cite{ye2023ip}. Next, we feed $z^c_0$ and $z^s_0$ into the Stable Diffusion U-Net and extract features from the third upsampling block of the U-Net, represented as $F_0^c \in\mathbb{R}^{HW\times C}$ and $F_0^s\in\mathbb{R}^{HW\times C}$ for images $z^c_0$ and $z^s_0$ (H and W represent the feature spatial
sizes, while C is the number of channels), respectively.
Then, the element of $i_{th}$ row and $j_{th}$ column in the correlation matrix is computed as:
\begin{equation}
	\mathcal{M}(i,j) = \frac{(\emph{F}_0^c(i)-\mu_{\emph{F}_0^c(i)})\cdot(\emph{F}_0^s(j)-\mu_{\emph{F}_0^s(j)})}{{\Vert\emph{F}_0^c(i)-\mu_{\emph{F}_0^c(i)}\Vert_2\;\Vert F_0^s}(j)-\mu_{\emph{F}_0^s(j)}\Vert_2}, \label{eq1}
\end{equation}
where $\mu_{\emph{F}_0^c}\in\mathbb{R}^{HW}$ and $\mu_{\emph{F}_0^s}\in\mathbb{R}^{HW}$ are mean vectors of $\emph{F}_0^c$ and $\emph{F}_0^s$ along the channel dimension. 

\begin{figure}[t]
	\centering
	\captionsetup[subfloat]{labelsep=none,format=plain,labelformat=empty}
    \includegraphics[width=1.0\linewidth]{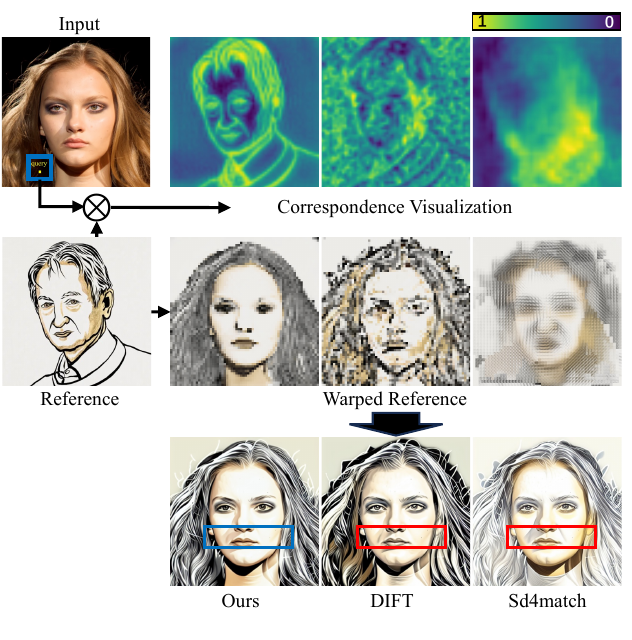}\\
    \vspace{-6mm}
    \caption{\textbf{Effect of our semantic correspondence.} For a query point (yellow point in input), the first row shows the similarity map computed by different methods. The second row displays the warped references. The third row shows the outputs generated using these different warped references. As seen, our correspondence achieves higher similarity in the semantic region of the reference, demonstrating its superior accuracy.}
\label{fig:ablation-correspondence}
  \vspace{-2mm}
\end{figure}

\vspace{0.5em}
\noindent \textbf{Semantic-Aware Image Warping.} With the semantic correspondence, we perform a differentiable warping of the reference portrait $z^s_0$ toward the input $z^c_0$, resulting in $z^{s\_w}_0$. Each pixel of $z^{s\_w}_0$ is computed as a weighted average of $z^s_0$:
\begin{equation}
    z^{s\_{w}}_0(i) = \sum_{j}\mathop{\rm{softmax}}\limits_{j} (\mathcal{M}(i,j)/\tau)\cdot z^s_0(j),  \label{eq2} 
\end{equation}
where $\tau$ is set to 0.01 to make the softmax curve sharper. To further enhance the accuracy of semantic correspondence, we introduce a mask warping loss during training, defined as:
\begin{equation}
	\mathcal{L}_{mask} = \Vert  M^c - M^{s\_w} \Vert_1,
\end{equation}
where $M^c$ represents the semantic mask of $z^c_0$, and $M^{s\_w}$ is obtained by warping the mask $M^s$ of $z^s_0$ following Equation \ref{eq2}. 
Besides, we introduce a cyclic warping consistency loss to reinforce style preservation during the warping process:
\begin{equation}
	\mathcal{L}_{cwc} = \mathcal{L}_{LPIPS}(z^s_0, z^{s^{\prime}\_w}),
\end{equation}
where $z^{s^{\prime}\_w}$ is obtained by warping $z^{s\_w}$ using the transposed correlation matrix $\mathcal{M^\top}$.  
By incorporating these two losses, the dense correspondence learned from diffusion models is further refined and enhanced. Figure \ref{fig:ablation-correspondence} compares the effect of replacing the correspondence in our method with two alternatives \cite{tang2023emergent}. As shown in the warped reference, output, and visualized correspondence map, the correspondence in our method is more accurate. This accuracy enables our method to generate portrait images that faithfully replicate the reference style while preserving the portrait's identity and facial structure details. 

\subsection{Dual Conditional Diffusion Model\label{ContentGuide}}
The proposed dual-conditional diffusion model leverages both structure guidance and style guidance to generate realistic portrait images. This novel architectural design enables the integration of high-frequency structures and semantically aligned style references, enhancing the quality and coherence of the generated images. 

\vspace{0.5em}
\noindent \textbf{Structure Guidance.} Unlike the previous method \cite{wang2024instantstyleplus}, our model utilizes the high-frequency information from the content image $z^c_0$ through ControlNet to provide style-independent structural guidance.
To achieve this, we apply the Haar Discrete Wavelet Transform (DWT) to extract the high-frequency components of the content image. Specifically, the Haar wavelet transform performs a depth-wise convolution with four kernels, \{LL$^\top$ LH$^\top$ HL$^\top$ HH$^\top$\}, where $\text{L} = \frac{1}{\sqrt{2}}[\,1\quad1\,]^\top$, $ \text{H} = \frac{1}{\sqrt{2}}[\,-1\quad1\,]^\top.$ Among these, LL$^\top$ acts as a low-pass filter, while LH$^\top$, HL$^\top$, and HH$^\top$\ serve as high-pass filters that capture high-frequency information in the horizontal, vertical, and diagonal directions, respectively. We use the outputs from the three high-pass filters LH$^\top$, HL$^\top$ and HH$^\top$ as the corresponding input to ControlNet, which can be formulated as:
\begin{align}
	c^{cnt} &= [z^c_{LH},z^c_{HL},z^c_{HH}] \\ 
    &= \text{Conv}([\text{LH}^\top, \text{HL}^\top, \text{HH}^\top],z^c_0). \label{LH}
\end{align}
Note that the original spatial size of $c^{cnt}$ is half that of $z^c_0$. Therefore, we upsample $c^{cnt}$ using bilinear interpolation and use the upsampled $c^{cnt}$ as input to the ControlNet. 

\vspace{0.5em}
\noindent \textbf{Style Guidance.} 
We employ a style adapter consisting of a text encoder and an image encoder \cite{ye2023ip}, integrated into the cross-attention layers for style guidance extraction and injection. The text prompt is fixed as ``a photo of a portrait," while image features is extracted from the penultimate layer of CLIP. The extracted latent representation is then passed through a projection layer to obtain the style guidance $c^{sty} = [c^t, c^i]$,  where $c^t$ represents the text feature and $c^i$ represents the image feature. These features are then fused with the main Denoising U-Net. Specifically, we perform decoupled cross-attention operations on both text and image features, injecting them into the corresponding layers of the Denoising U-Net. The decoupled cross-attention mechanism can be illustrated as:
\begin{align}
Z^{new}&={\rm softmax}(\frac{QK^t}{\sqrt{d}})V^t+ \lambda \cdot{\rm softmax}(\frac{QK^i}{\sqrt{d}})V^i,
\end{align}
where $\lambda$ is a weight that balances the contributions of text and image prompts. $Q$, $K^t$, $V^t$ represent the query, key, and value matrices for text cross-attention, while $K^i$ and $V^i$ correspond to image cross-attention. Given the query features $Z$ from the main denoising UNet, we define $Q = ZW_q$, $K^i = c^iW^i_k$, $V^i =  c^iW_v^i$, $K^t = c^tW^t_k$ and $V^t =  c^tW_v^t$, where only $W^i_k$ and $W^i_v$ are trainable weights.

\subsection{Initial Latent AdaIN-Wavelet Transform\label{Initlatent}}
In portrait style transfer, color tone plays a crucial role in conveying style information. If sampling begins from the latent $z_T^c$ of the content image, the generated image will largely retain its original color tone. To achieve an initial latent representation that preserves the structural details of the input image while enhancing local color style transfer, we propose a novel AdaIN-Wavelet transform for latent initialization. First, we compute the latent representation of the warped reference image $z_0^{s\_w}$ using DDIM inversion, denoted as $z_T^{s\_w}$. Since initiating the denoising process from $z_T^{s\_w}$ strongly enhances color transfer, directly using it may result in a blurred output. To address this, we propose blending the low-frequency components of $z_T^{s\_w}$ with the high-frequency components of the content latent, striking a balance between effective stylization and content detail preservation. We first apply AdaIN between $z_T^{s\_w}$ and $z_T^{c}$, formulated as:
\begin{equation}
    z_{T}^{cs^{\prime}} = \sigma(z_T^{s\_w})(\frac{z_T^{c}-\mu(z_T^{c})}{\sigma(z_T^{c})})
    +\mu(z_T^{s\_w}),
\end{equation}
where $\sigma()$ and $\mu()$ denote channel-wise mean and standard
deviation, respectively. We then apply the wavelet transform to integrate the high-frequency information from $z_{T}^{cs^{\prime}}$ and the low-frequency informatoin from $z_T^{s\_w}$, which is obtained as: 
\begin{equation}
    z_{T}^{cs} = \text{IDWT}(z_{T,LL}^{s\_w}, z_{T,LH}^{cs^{\prime}},z_{T,HL}^{cs^{\prime}},z_{T,HH}^{cs^{\prime}}),
\end{equation}
where IDWT denotes  the inverse wavelet transform.

To control the strength of stylization, we introduce a parameter $\gamma$. Since stylization is influenced by both the latent initialization and the style adapter, $\gamma$ is applied in two key areas. First, we interpolate between the input latent $z_T^c$ and stylized latent $z_{T}^{cs}$:
\begin{equation}
    z_{T}^{cs} = \gamma \times z_{T}^{cs} + (1-\gamma) \times z_{T}^{c}.
\end{equation}
Similarly, $\gamma$ is used in style adapter to interpolate between the feature embeddings of the warped reference image and the input image. In this paper, we set $\gamma=1.0$ by default for best stylization effects. 

\begin{figure*}[t]
    \centering
    \captionsetup[subfloat]{labelsep=none,format=plain,labelformat=empty}
    \begin{subfigure}[c]{0.12\textwidth}
        \includegraphics[width=1.0\textwidth]
        {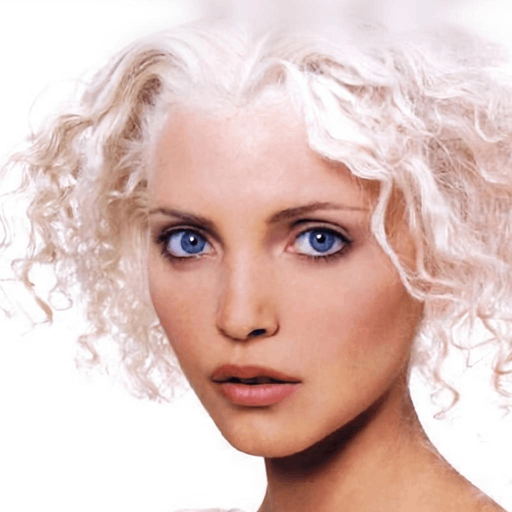}
        \caption{Reference}
    \end{subfigure}
    \begin{subfigure}[c]{0.12\textwidth}
        \includegraphics[width=1.0\textwidth]
        {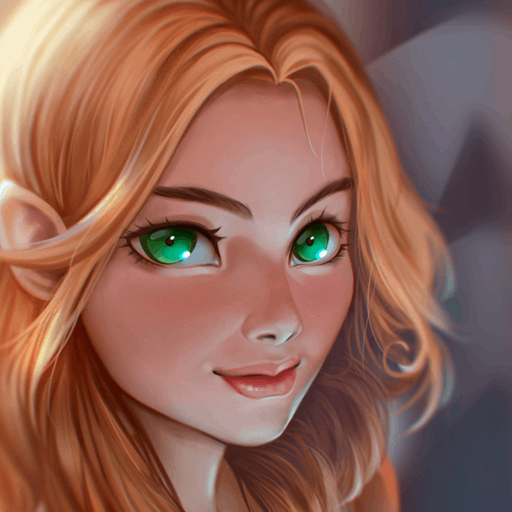}
        \caption{Input}
    \end{subfigure}
    \begin{subfigure}[c]{0.12\textwidth}
        \includegraphics[width=1.0\textwidth]{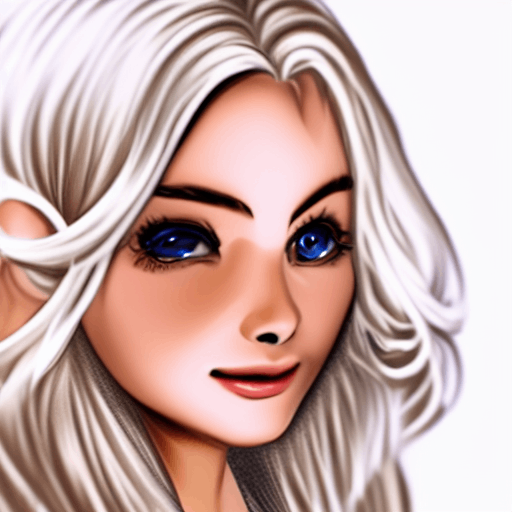}
        \caption{w/o Cont.}
    \end{subfigure}
    \begin{subfigure}[c]{0.12\textwidth}
        \includegraphics[width=1.0\textwidth]{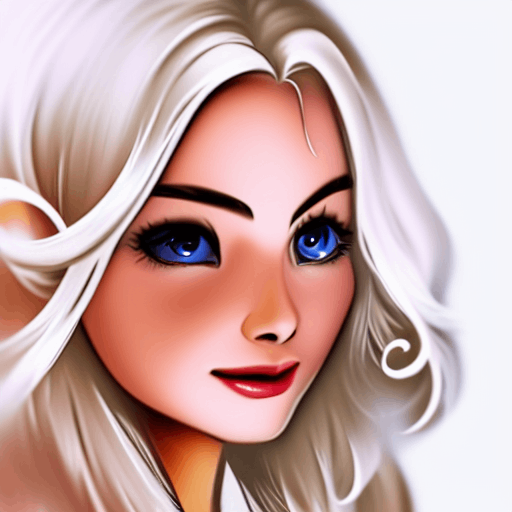}
        \caption{Canny Cont.}
    \end{subfigure}
    \begin{subfigure}[c]{0.12\textwidth}
        \includegraphics[width=1.0\textwidth]{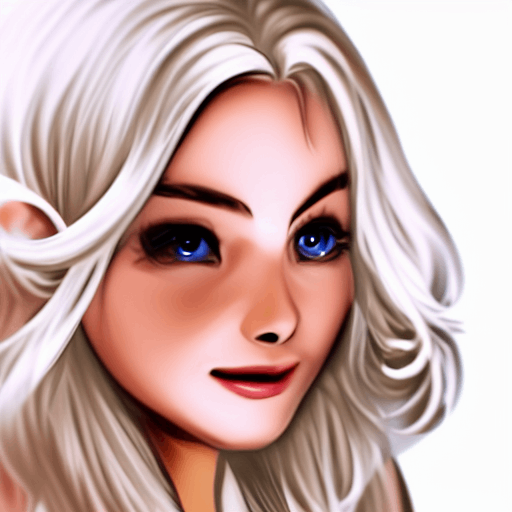}
        \caption{Input Cont.}
    \end{subfigure}
    \begin{subfigure}[c]{0.12\textwidth}
        \includegraphics[width=1.0\textwidth]{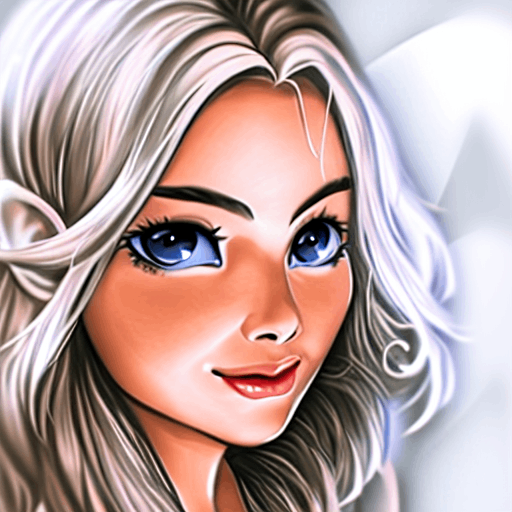}
        \caption{w/o style adapter}
    \end{subfigure}
    \begin{subfigure}[c]{0.12\textwidth}
        \includegraphics[width=1.0\textwidth]{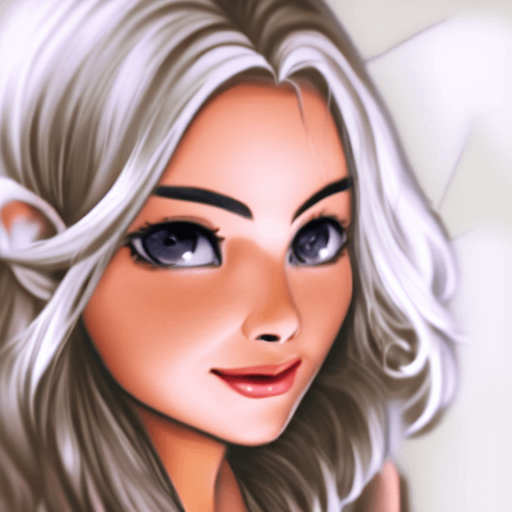}
        \caption{IP-A.(4 tokens)}
    \end{subfigure}
    \begin{subfigure}[c]{0.12\textwidth}
        \includegraphics[width=1.0\textwidth]
        {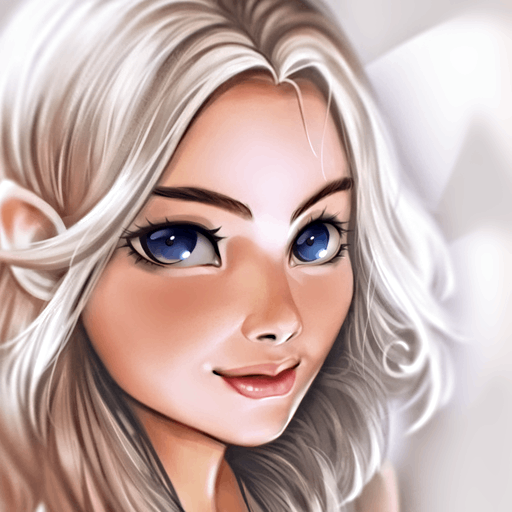}
        \caption{Ours}
    \end{subfigure}
    \vspace{-2mm}
    \caption{\textbf{Effect of the ControlNet and style adapter.} }
	\label{fig:ablation-ControlNet}
\end{figure*}	

\begin{figure*}[t]
	\centering
	\captionsetup[subfloat]{labelsep=none,format=plain,labelformat=empty}
    \begin{subfigure}[c]{0.15\textwidth}
        \includegraphics[width=1.0\textwidth]{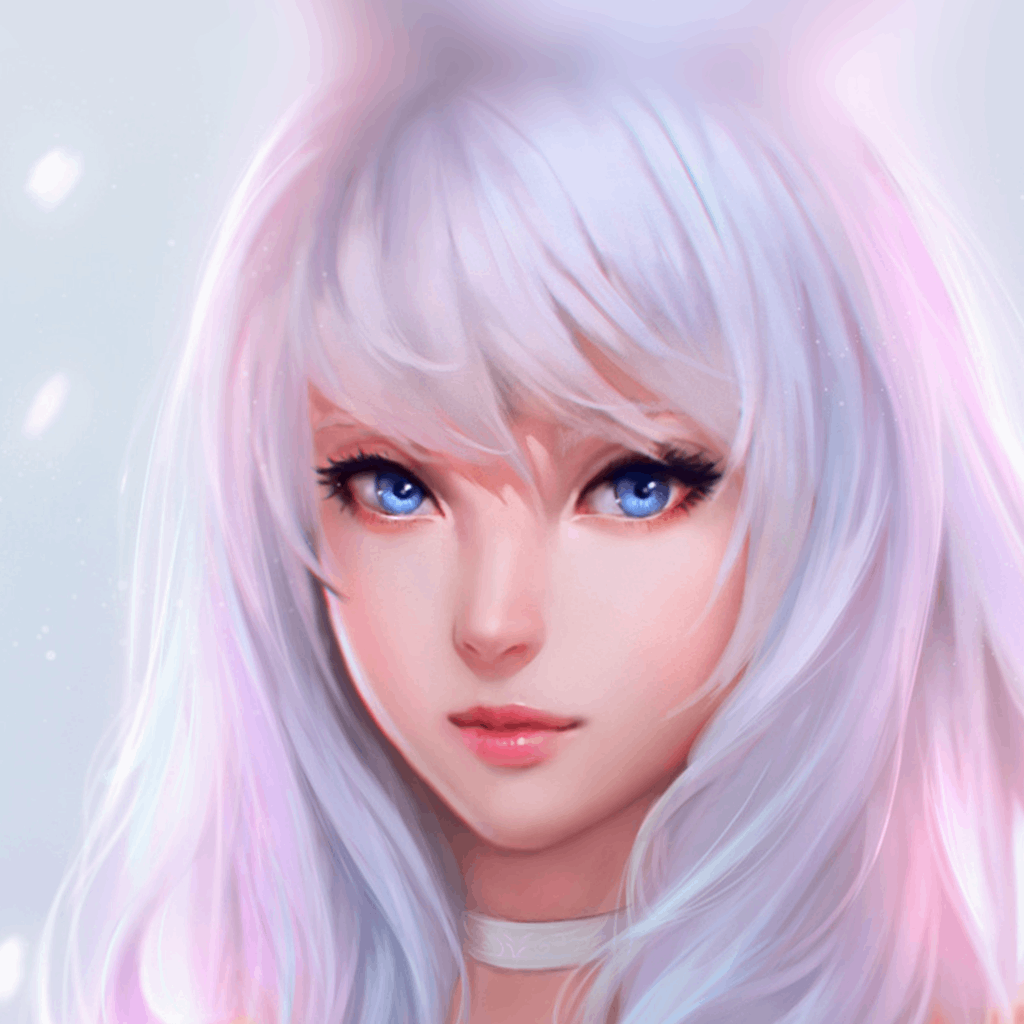}\vspace{1.2pt}
        \caption{Reference}
    \end{subfigure}
    \begin{subfigure}[c]{0.15\textwidth}
        \includegraphics[width=1.0\textwidth]{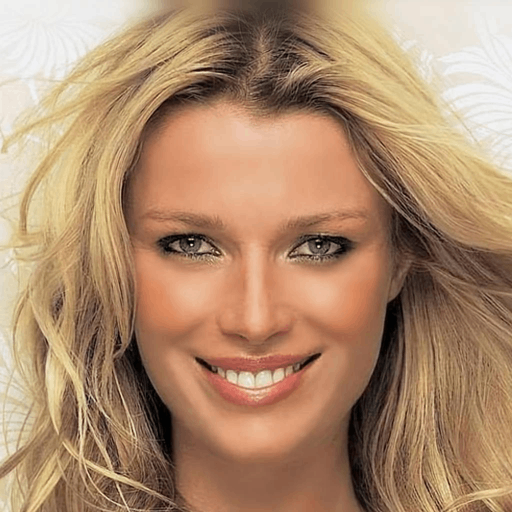}\vspace{1.2pt}
        \caption{Input}
    \end{subfigure}
    \begin{subfigure}[c]{0.15\textwidth}
        \includegraphics[width=1.0\textwidth]{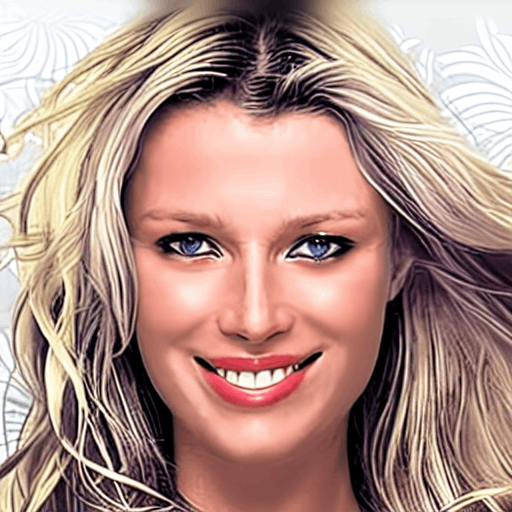}\vspace{1.2pt}
        \caption{Input latent ($z_T^c$)}
    \end{subfigure}
    \begin{subfigure}[c]{0.15\textwidth}
        \includegraphics[width=1.0\textwidth]{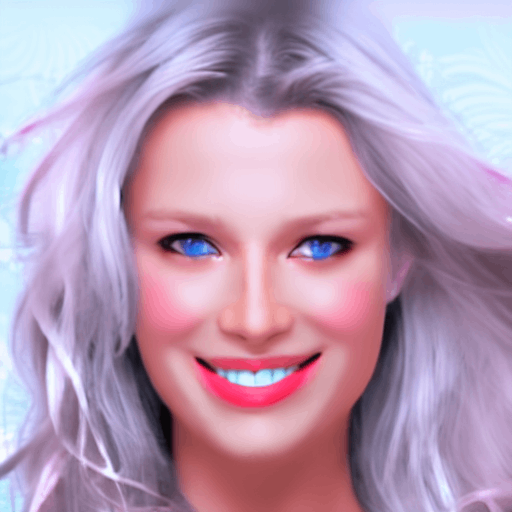}\vspace{1.2pt}
        \caption{w.r. latent ($z_T^{s\_w}$)}
    \end{subfigure}
    \begin{subfigure}[c]{0.15\textwidth}
        \includegraphics[width=1.0\textwidth]{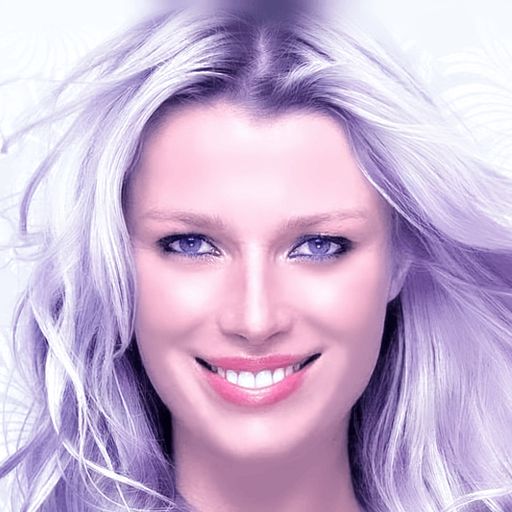}\vspace{1pt}
        \caption{AdaIN ($z_{T}^{cs^{\prime}}$)}
    \end{subfigure}
    \begin{subfigure}[c]{0.15\textwidth}
        \includegraphics[width=1.0\textwidth]{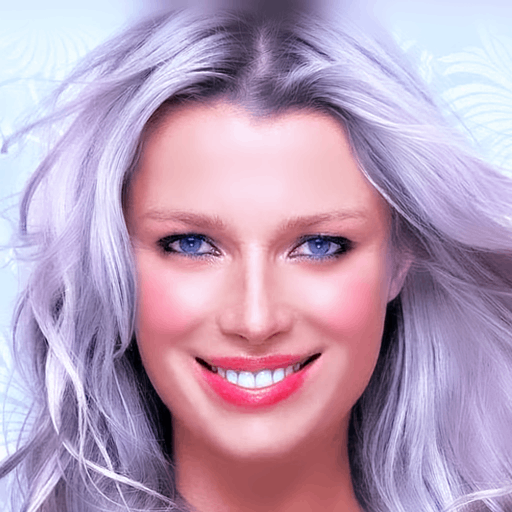}\vspace{1.2pt}
        \caption{Ours ($z_{T}^{cs}$)}
    \end{subfigure}
    \vspace{-2mm}
	\caption{\textbf{Effect of the initialization of latent noise.} ``w.r." stands for warped reference. }
	\label{fig:ablation-noise}
    \vspace{-2mm}
\end{figure*}	
\begin{figure}[t]
    \centering
    \captionsetup[subfloat]{labelsep=none,format=plain,labelformat=empty}
    \begin{subfigure}[c]{0.09\textwidth}
        \includegraphics[width=1.0\textwidth]
        {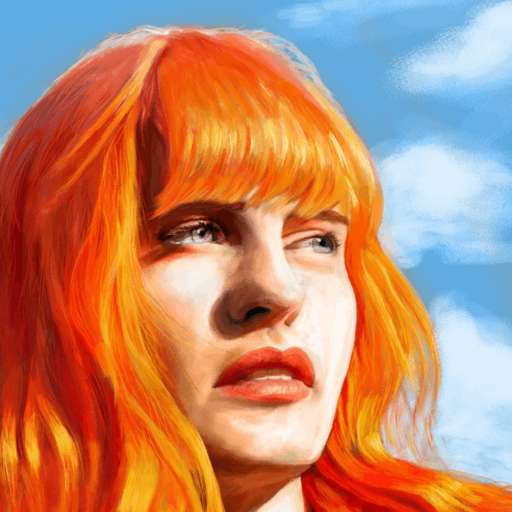}\vspace{1pt}
        \caption{Reference}
    \end{subfigure}
    \begin{subfigure}[c]{0.09\textwidth}
        \includegraphics[width=1.0\textwidth]
        {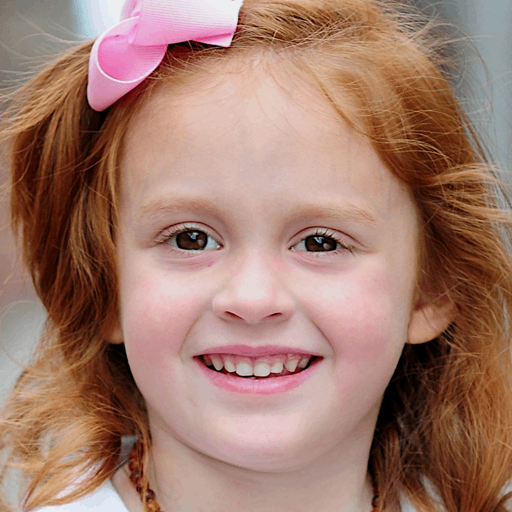}\vspace{1pt}
        \caption{Input}
    \end{subfigure}
    \begin{subfigure}[c]{0.09\textwidth}
        \includegraphics[width=1.0\textwidth]{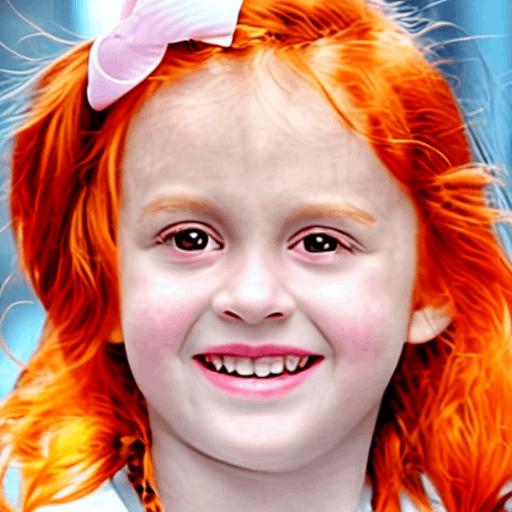}\vspace{1pt}
        \caption{Hair}
    \end{subfigure}
    \begin{subfigure}[c]{0.09\textwidth}
        \includegraphics[width=1.0\textwidth]{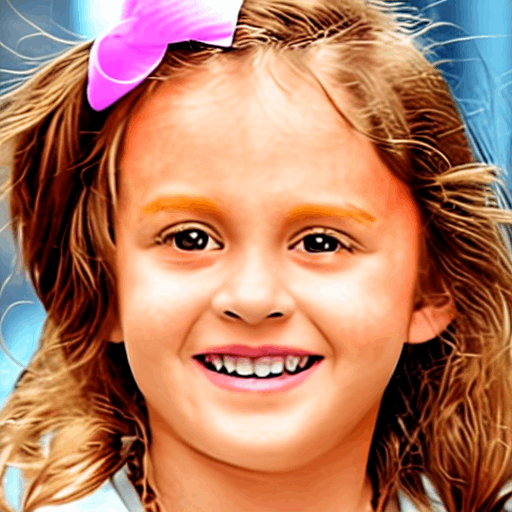}\vspace{1pt}
        \caption{face}
    \end{subfigure}
    \begin{subfigure}[c]{0.09\textwidth}
        \includegraphics[width=1.0\textwidth]{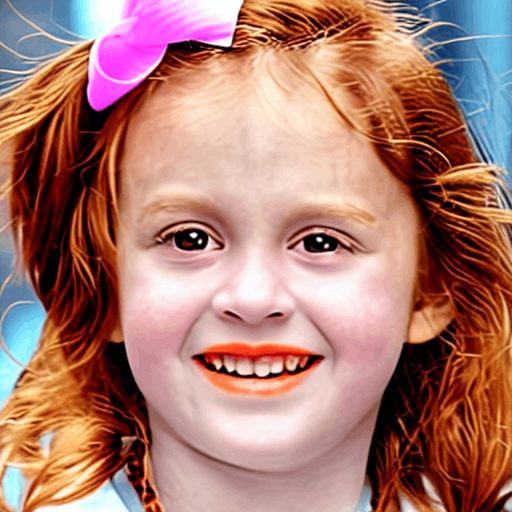}\vspace{1pt}
        \caption{lips}
    \end{subfigure}
    \vspace{-2mm}
    \caption{\textbf{Controllable region-specific style transfer.}}
	\label{fig:regionst}
\end{figure}

\begin{table}[t]
	\centering
    \scalebox{0.94}{
	\begin{tabular}{lccc}
		\toprule
		{Method} & {Gram loss$\downarrow$} & {LPIPS$\downarrow$}&{ID$\downarrow$} \\
		\midrule
        {\citet{shih_sig14}} & 0.376 & 0.187 & 0.093  \\ 
        {\citet{chen2021deepfaceediting}} & 0.287 & 0.156 & 0.344  \\ 
        {\citet{zhou2024deformable}} & 0.688 & 0.393 & 0.657  \\ 
        {\citet{wang2025ppst}} &\textbf{0.208}  &0.181  & 0.106    \\ 
        {IP-A\cite{ye2023ip} + C.N.\cite{zhang2023adding}} & 2.835 & 0.245 & 0.774  \\ 
        {\citet{deng2023zzeroshotstyletransfer}} & 1.745 & 0.162 & 0.393  \\ 
        {StyleID\cite{chung2023style}} & 0.505 & 0.198 & 0.222  \\ 
        {InstantStyle + \cite{wang2024instantstyleplus}} & 0.557 & 0.294 & 0.272  \\ 
        \midrule
        {Ours} & 0.274 & \textbf{0.116} & \textbf{0.057}  \\
       {Ours (w/o ControlNet)} &\underline{0.236}  &0.333 & 0.450   \\ 
       {Ours (w/o style adapter)} &0.548  & \underline{0.145} & 0.086   \\ 
       {Ours (w/ Init AdaIN)} &1.196 &0.151 &\underline{0.062}    \\ 
		\bottomrule
	\end{tabular}}
    \caption{\textbf{Quantitative comparison on CelebAMask-HQ.}}
    \label{table:quan_comp}
 \vspace{-3mm}
\end{table}

\subsection{Training and Implementation Details \label{Training}}
We implement our method using Stable Diffusion 1.5 and follow a two-stage training strategy. 
In the first stage, we train the semantic adapter while freezing the other parameters of the pre-trained diffusion model. To ensure the features encode meaningful information, we incorporate a noise prediction loss during this stage:
\begin{equation}
    \label{noise-pred}
	\mathcal{L}_{sem} = \mathbb{E}_{z_t,t,I^x,\epsilon\sim\mathcal{N}(0,I)} \Vert \epsilon - \epsilon_\theta(z_t,t,I^x) \Vert_2^2,
\end{equation}
where $I^x$ represent the input image. The total loss at this stage is:
\begin{equation}
    \label{second}
	\mathcal{L} = \mathcal{L}_{sem} + \lambda_c\mathcal{L}_{cyc} + \lambda_m \mathcal{L}_{mask},
\end{equation}
where $\lambda_c$ and $\lambda_m$ are set to 1 and 10, respectively. 

In the second training stage, the ControlNet becomes trainable. We train the model to generate images conditioned on both the style and content inputs. The loss function at this stage can be formulated as follows:
\begin{equation}
    \label{noise-pred}
	\mathcal{L}_{rec} = \mathbb{E}_{z_t,t,c^{cnt},c^{sty},\epsilon\sim\mathcal{N}(0,I)} \Vert \epsilon - \epsilon_\theta(z_t,t,c^{cnt},c^{sty}) \Vert_2^2,
\end{equation}
where both $c^{cnt}$ and $c^{sty}$ are extracted from the input image as there are no image pairs available for our tasks. 

Our method is implemented in PyTorch \cite{paszke2019pytorch}. We utilize 2 Nvidia RTX 4090 GPUs with a batch size of 1 per GPU for training.
The AdamW \cite{loshchilov2017decoupled} optimizer is employed, with a learning rate of 1e-5. The first stage of training runs for 500K iterations, while the second stage is trained for 300K iterations. The training process requires approximately 24GB of memory, and testing necessitates 16GB of memory for images with a resolution of 512 $\times$ 512. For testing, we use the DDIM sampler \cite{liu2022pseudo} with 10 timesteps for inversion and 30 timesteps for sampling.

\begin{figure}[t]
    \centering
    \captionsetup[subfloat]{labelsep=none,format=plain,labelformat=empty}
    \begin{subfigure}[c]{0.075\textwidth}
        \includegraphics[width=1.0\textwidth]
        {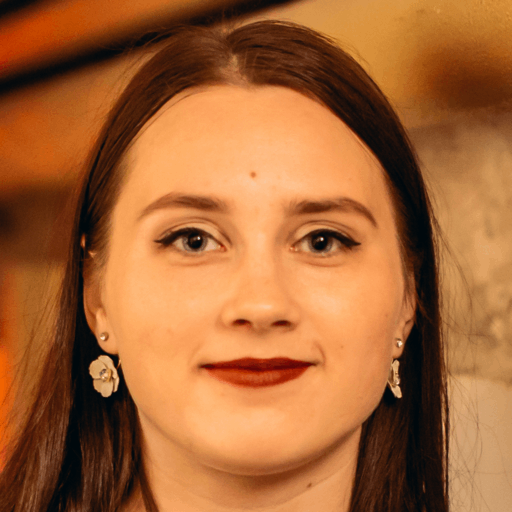}\vspace{1pt}
        \includegraphics[width=1.0\textwidth]
        {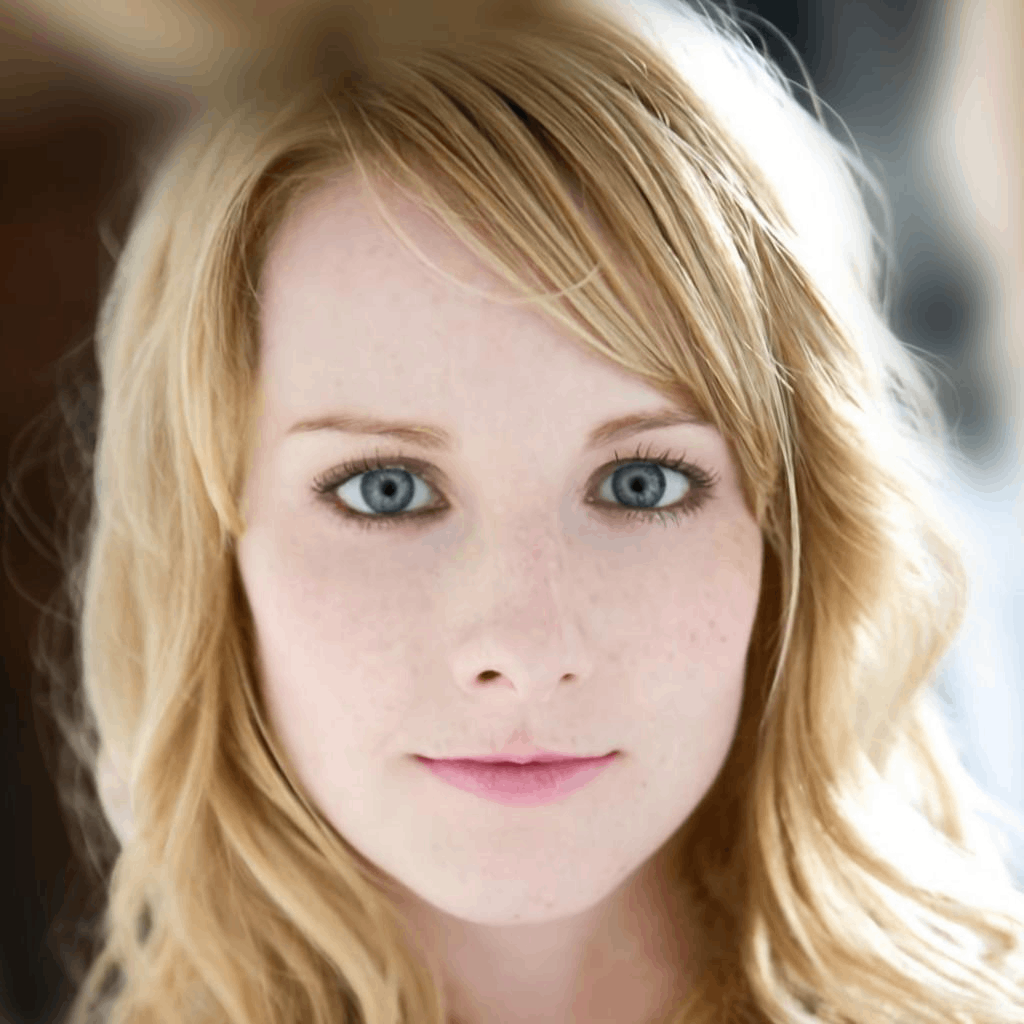}\vspace{1pt}
        \caption{Reference}
    \end{subfigure}
    \begin{subfigure}[c]{0.075\textwidth}
        \includegraphics[width=1.0\textwidth]
        {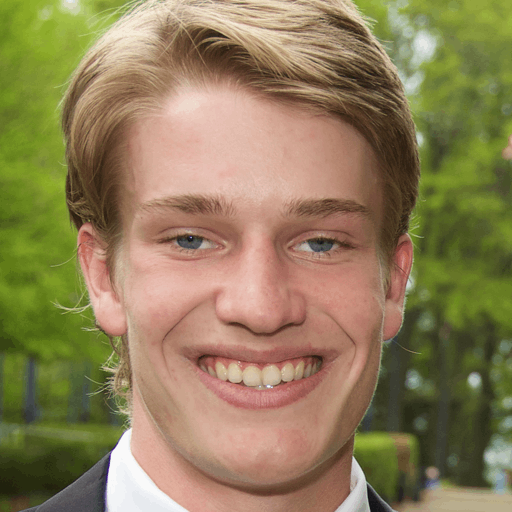}\vspace{1pt}
        \includegraphics[width=1.0\textwidth]
        {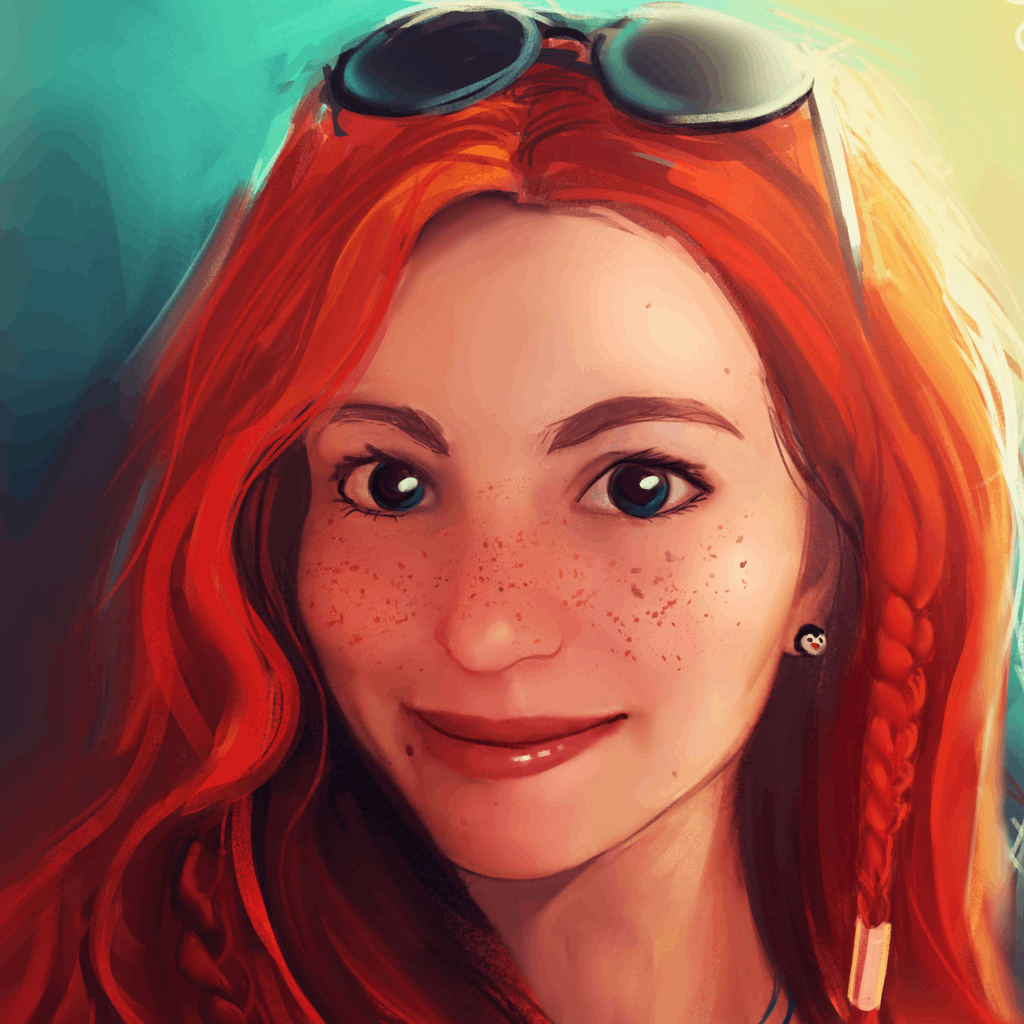}\vspace{1pt}
        \caption{Input}
    \end{subfigure}
    \begin{subfigure}[c]{0.075\textwidth}
        \includegraphics[width=1.0\textwidth]{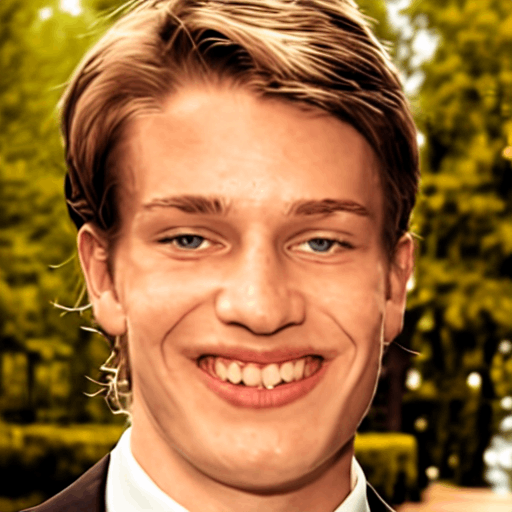}\vspace{1pt}
        \includegraphics[width=1.0\textwidth]{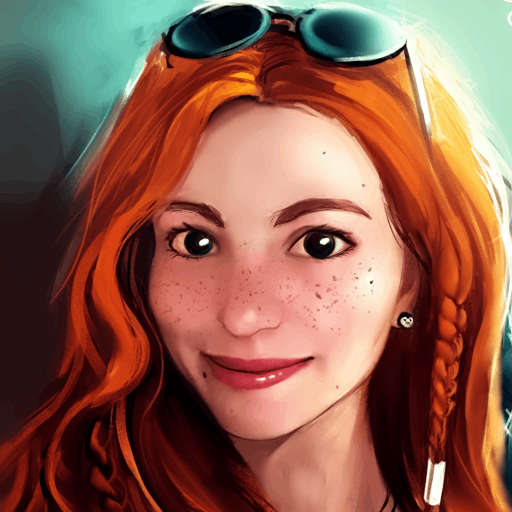}\vspace{1pt}
        \caption{$\gamma =0.25$}
    \end{subfigure}
    \begin{subfigure}[c]{0.075\textwidth}
        \includegraphics[width=1.0\textwidth]{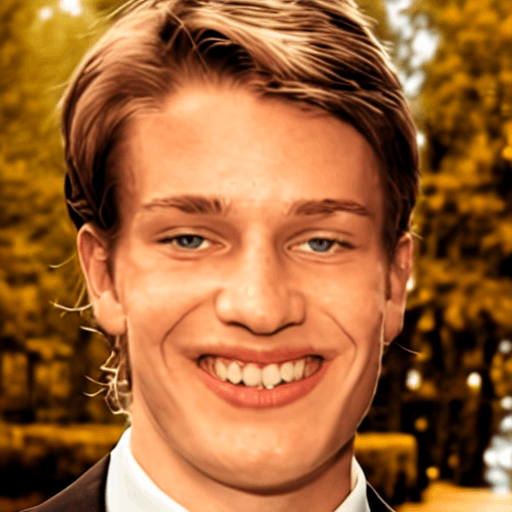}\vspace{1pt}
        \includegraphics[width=1.0\textwidth]{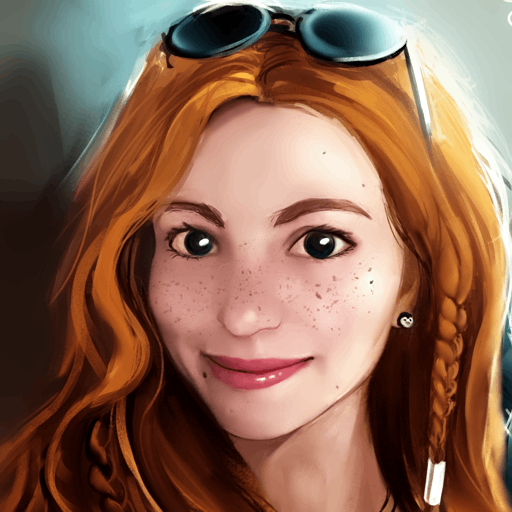}\vspace{1pt}
        \caption{$\gamma =0.50$}
    \end{subfigure}
    \begin{subfigure}[c]{0.075\textwidth}
        \includegraphics[width=1.0\textwidth]{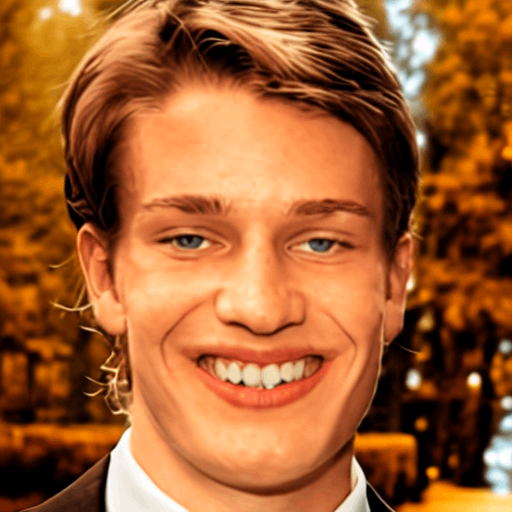}\vspace{1pt}
        \includegraphics[width=1.0\textwidth]{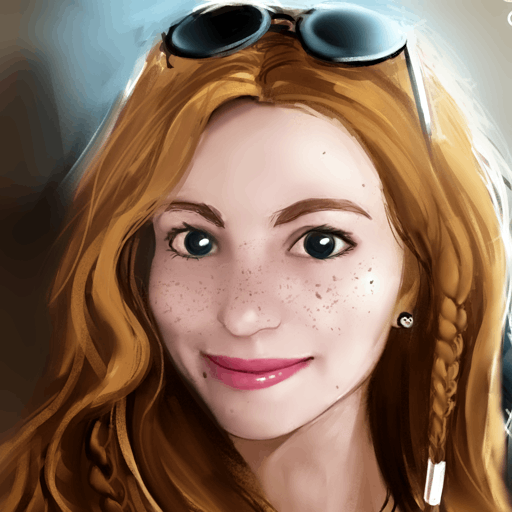}\vspace{1pt}
        \caption{$\gamma =0.75$}
    \end{subfigure}
    \begin{subfigure}[c]{0.075\textwidth}
        \includegraphics[width=1.0\textwidth]
{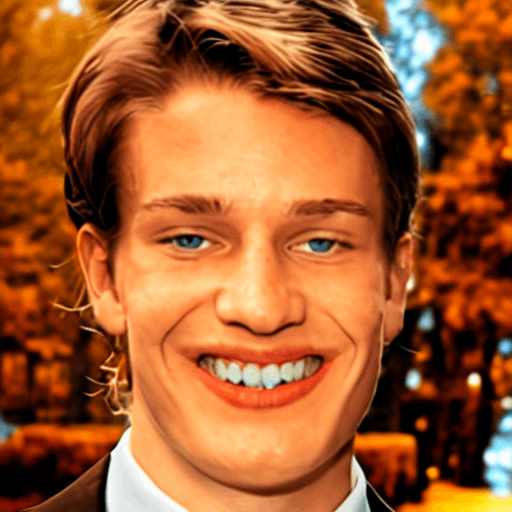}\vspace{1pt}
        \includegraphics[width=1.0\textwidth]
{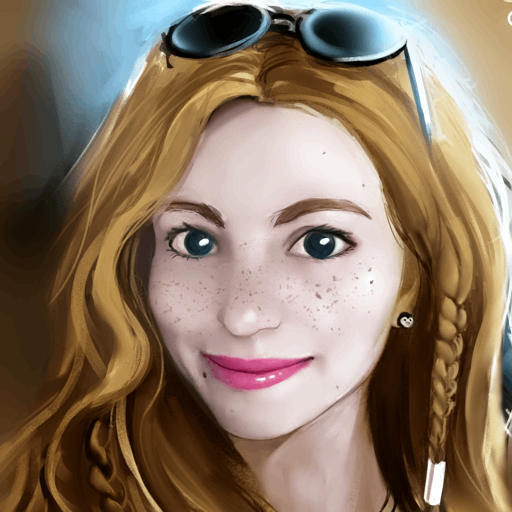}\vspace{1pt}
        \caption{$\gamma =1.0$}
    \end{subfigure}
    \vspace{-2mm}
    \caption{\textbf{Style interpolation.} }
	\label{fig:interpolation}
  \vspace{-3mm}
\end{figure}

\begin{table}[t]
	\centering
    \scalebox{0.99}{
	\begin{tabular}{lccc}
		\toprule
		{Method} & {Gram loss$\downarrow$} & {LPIPS$\downarrow$}&{ID$\downarrow$} \\
		\midrule
        {\citet{shih_sig14}}&0.802&0.156&0.105  \\ 
        {\citet{chen2021deepfaceediting}} &0.926 &0.169 &0.316 \\ 
        {\citet{zhou2024deformable}}&0.827&0.423&0.669  \\ 
        {\citet{wang2025ppst}}&1.488&\underline{0.119}&\underline{0.096}  \\ 
        {IP-A\cite{ye2023ip} + C.N.\cite{zhang2023adding}}&4.211&0.375&0.763  \\ 
        {\citet{deng2023zzeroshotstyletransfer}}&1.493 &0.178&0.331  \\ 
        {StyleID\cite{chung2023style}}&1.343&0.149&0.165 \\ 
        {InstantStyle+ \cite{wang2024instantstyleplus}}&\underline{0.723}&0.192&0.203  \\ 
        \midrule
        {Ours} &\textbf{0.657}&\textbf{0.083}&\textbf{0.087} \\
		\bottomrule
	\end{tabular}}
    \caption{\textbf{Quantitative comparison on a mixed dataset with portraits from multiple domains.}}
    \label{table:quan_comp2}
 \vspace{-3mm}
\end{table}

\section{Experiments}
\subsection{Datasets and Evaluation Metrics} 
\noindent \textbf{Datasets.} We evaluate our method on CelebAMask-HQ \cite{CelebAMask-HQ}, FFHQ \cite{StyleGAN}, and AAHQ \cite{liu2021blendgan}. CelebAMask-HQ contains 30K high-resolution portrait images, each paired with a facial segmentation mask covering 19 facial semantic categories. FFHQ consists of 70K high-quality images depicting a wide range of human faces. AAHQ includes approximately 25K high-quality artistic images with varied painting styles, color tones, and face attributes. To demonstrate the generalization ability of our method, we train it solely on the 28K training images from the CelebAMask-HQ dataset \cite{wang2025ppst}.

\vspace{0.5em}
\noindent \textbf{Metrics.} To evaluate the stylization effects for semantically corresponding regions, we follow \citet{wang2025ppst} to compute the Gram loss between the input image and the warped reference. Additionally, we use LPIPS \cite{Zhang_2018_CVPR} to measure content preservation performance and adopt the Identity Distance (ID) \cite{deng2019arcface}, as used in \cite{li2023w-plus-adapter, mu2022coordgan}, to quantitatively assess identity preservation.

\subsection{Comparison with State-of-the-art Methods} 

\noindent \textbf{Baselines.} We compare our method with two types of state-of-the-art style transfer methods: (i) conventional portrait style transfer methods, including Shih et al.\cite{shih_sig14}, \citet{chen2021deepfaceediting}, \citet{zhou2024deformable}, and \citet{wang2025ppst}, and (ii) diffusion-based style transfer methods, including IP-Adapter (IP-A) \cite{ye2023ip} + ControlNet (C.N.) \cite{zhang2023adding}, \citet{deng2023zzeroshotstyletransfer}, StyleID \cite{chung2023style}, and InstantStyle-Plus (InstantStyle+) \cite{wang2024instantstyleplus}. For learning-based methods, we retrain them on CelebAMask-HQ using the publicly available training code provided by the authors, with the recommended parameter settings. For methods with pre-trained models only, we fine-tune their models on CelebAMask-HQ.

\begin{figure*}[t]
	\captionsetup[subfloat]{labelsep=none,format=plain,labelformat=empty}
    \begin{subfigure}[c]{0.163\textwidth}
        \includegraphics[width=1.0\textwidth]{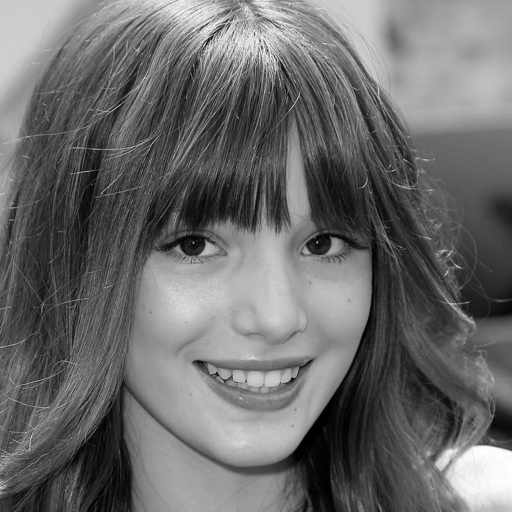}\vspace{1pt}
        \caption{Input}
    \end{subfigure}
    \begin{subfigure}[c]{0.163\textwidth}
        \includegraphics[width=1.0\textwidth]{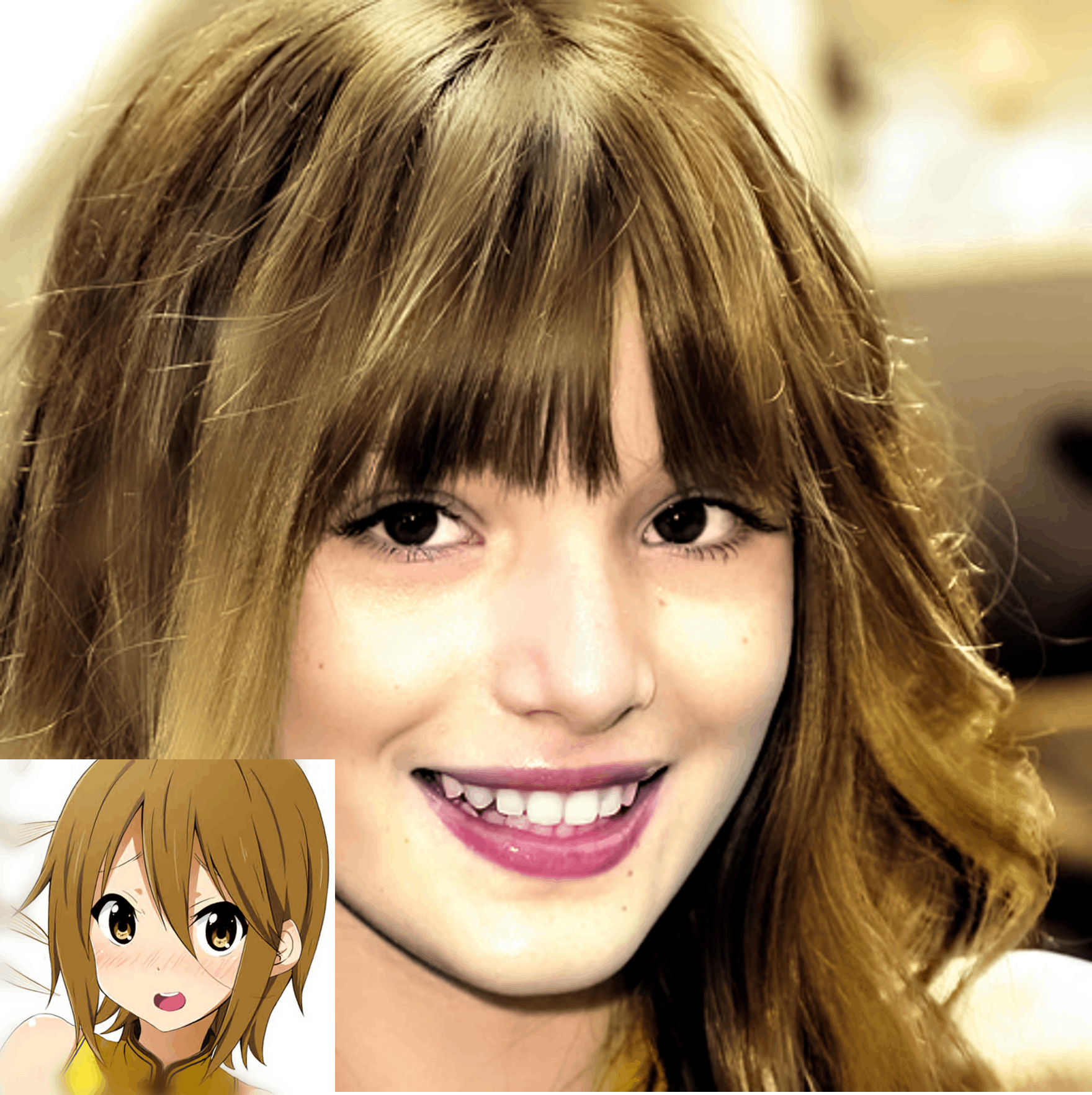}\vspace{1pt}
        \caption{Output}
    \end{subfigure}
    \begin{subfigure}[c]{0.163\textwidth}
        \includegraphics[width=1.0\textwidth]{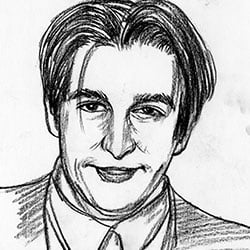}\vspace{1pt}
        \caption{Input}
    \end{subfigure}
    \begin{subfigure}[c]{0.163\textwidth}
        \includegraphics[width=1.0\textwidth]{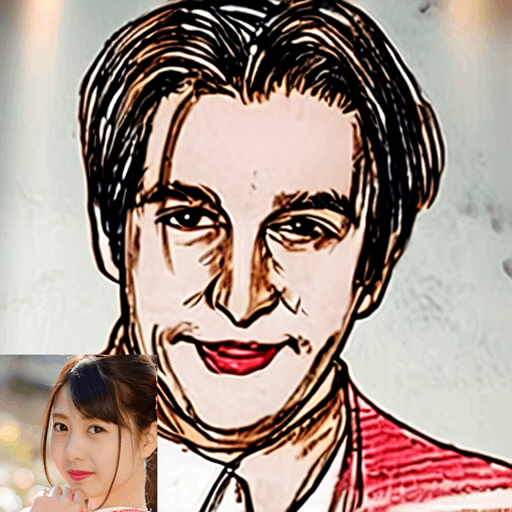}\vspace{1pt}
        \caption{Output}
    \end{subfigure}
    \begin{subfigure}[c]{0.163\textwidth}
        \includegraphics[width=1.0\textwidth]{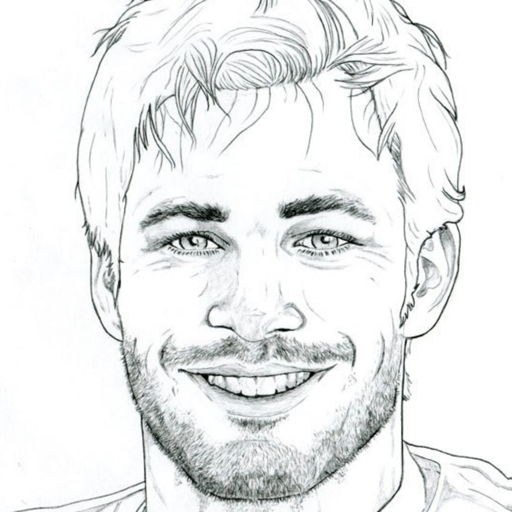}\vspace{1pt}
        \caption{Input}
    \end{subfigure}
    \begin{subfigure}[c]{0.163\textwidth}
        \includegraphics[width=1.0\textwidth]{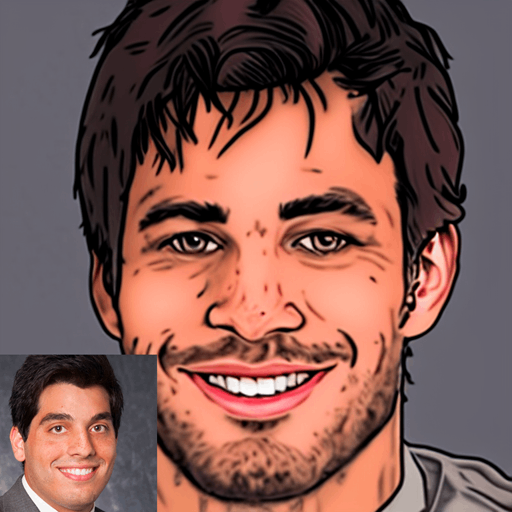}\vspace{1pt}
        \caption{Output}
    \end{subfigure}
     \vspace{-2mm}
	\caption{\textbf{Results of
adding color to grays-scale and sketch portraits}. }
	\label{fig:color}

 
\end{figure*}
\begin{figure*}[t]
	\captionsetup[subfloat]{labelsep=none,format=plain,labelformat=empty}
    \begin{subfigure}[c]{0.163\textwidth}
        \includegraphics[width=1.0\textwidth]{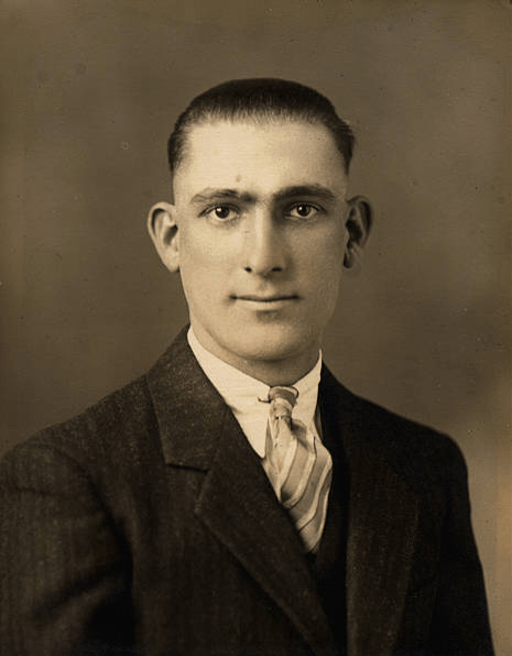}\vspace{1pt}
        \caption{Input}
    \end{subfigure}
    \begin{subfigure}[c]{0.163\textwidth}
        \includegraphics[width=1.0\textwidth]{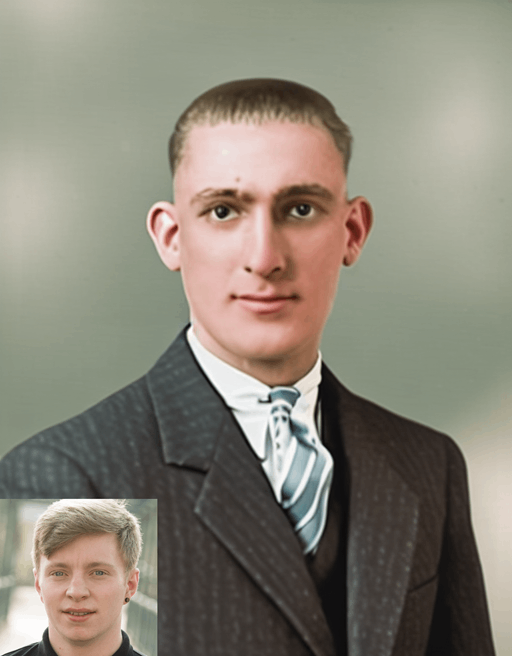}\vspace{1pt}
        \caption{Output}
    \end{subfigure}
    \begin{subfigure}[c]{0.163\textwidth}
        \includegraphics[width=1.0\textwidth]{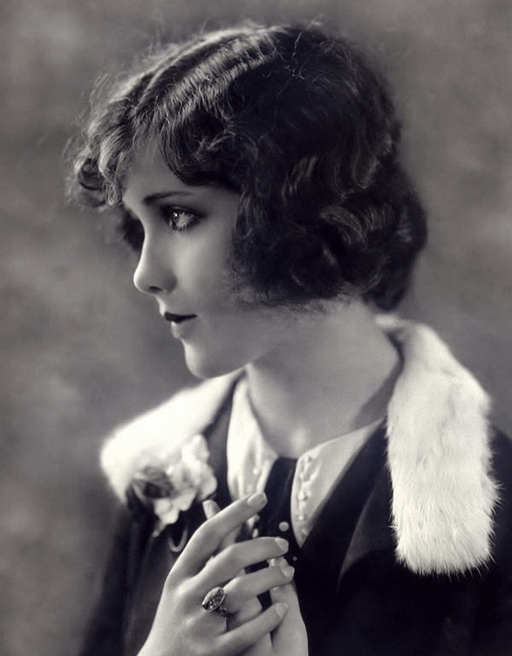}\vspace{1pt}
        \caption{Input}
    \end{subfigure}
    \begin{subfigure}[c]{0.163\textwidth}
        \includegraphics[width=1.0\textwidth]{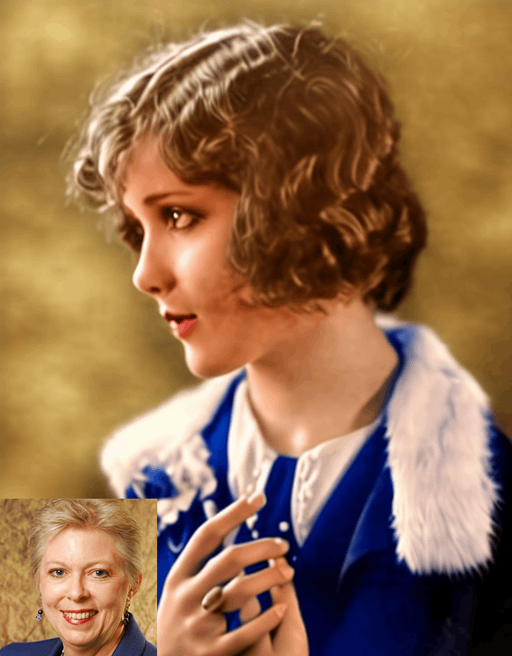}\vspace{1pt}
        \caption{Output}
    \end{subfigure}
    \begin{subfigure}[c]{0.163\textwidth}
        \includegraphics[width=1.0\textwidth]{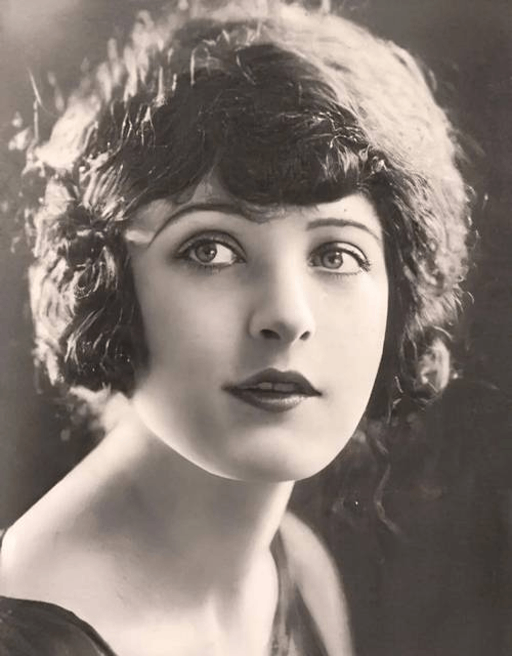}\vspace{1pt}
        \caption{Input}
    \end{subfigure}
    \begin{subfigure}[c]{0.163\textwidth}
        \includegraphics[width=1.0\textwidth]{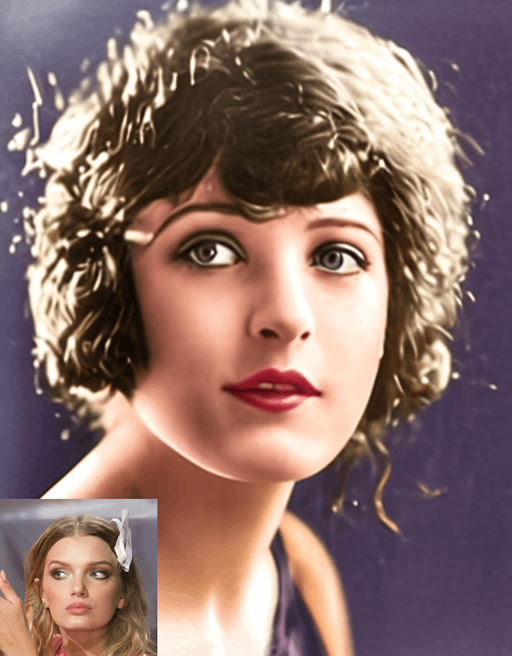}\vspace{1pt}
        \caption{Output}
    \end{subfigure}
     \vspace{-2mm}
	\caption{\textbf{Results of modernizing old photographs.}}
	\label{fig:oldphotos}
\vspace{-2mm}
\end{figure*}	

\begin{figure}[t]
	\centering
	\captionsetup[subfloat]{labelsep=none,format=plain,labelformat=empty}
    \begin{subfigure}[c]{0.115\textwidth}
        \includegraphics[width=1.0\textwidth]{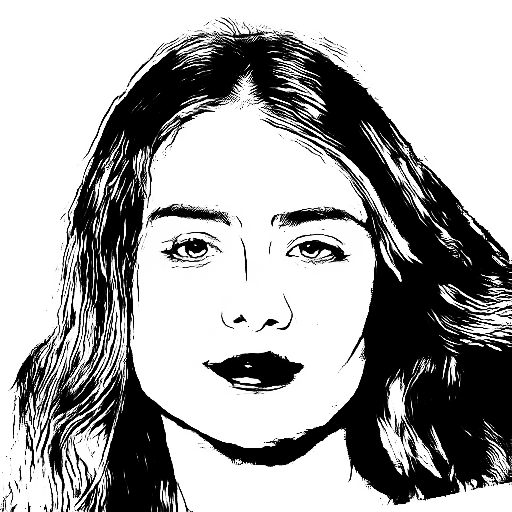}\vspace{1pt}
        \includegraphics[width=1.0\textwidth]
        {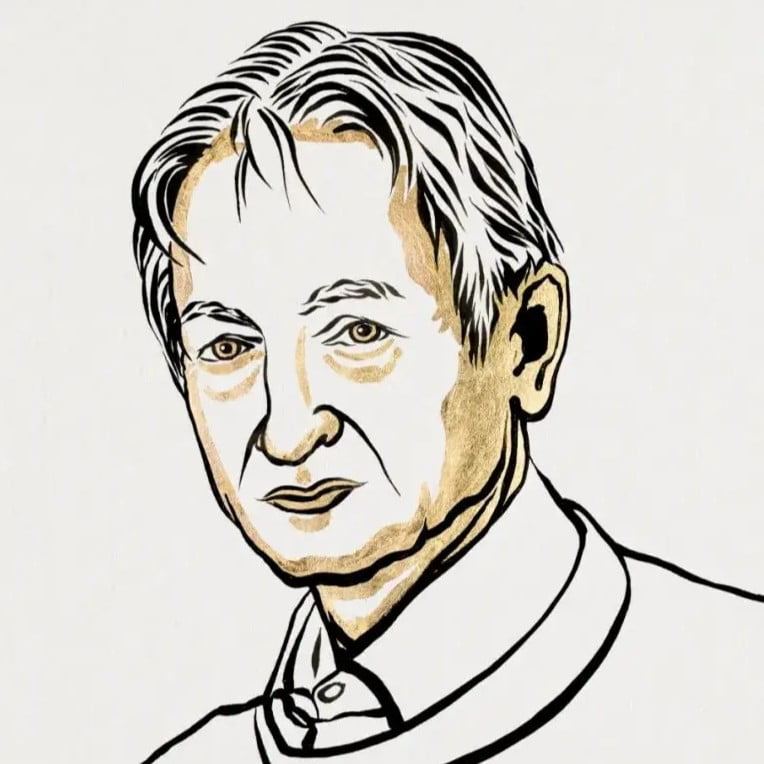}\vspace{1pt}
        \caption{Reference}
    \end{subfigure}
    \begin{subfigure}[c]{0.115\textwidth}
        \includegraphics[width=1.0\textwidth]{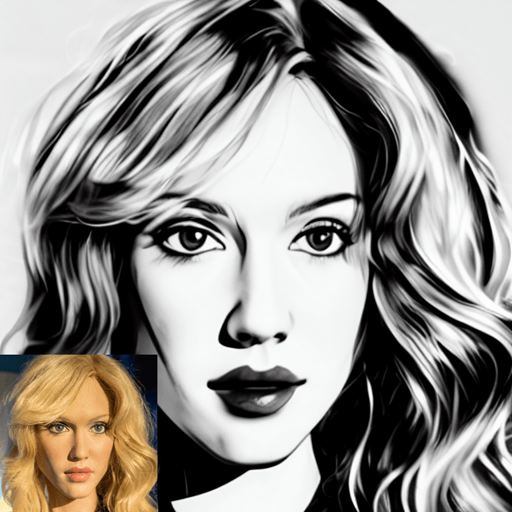}\vspace{1pt}
        \includegraphics[width=1.0\textwidth]
        {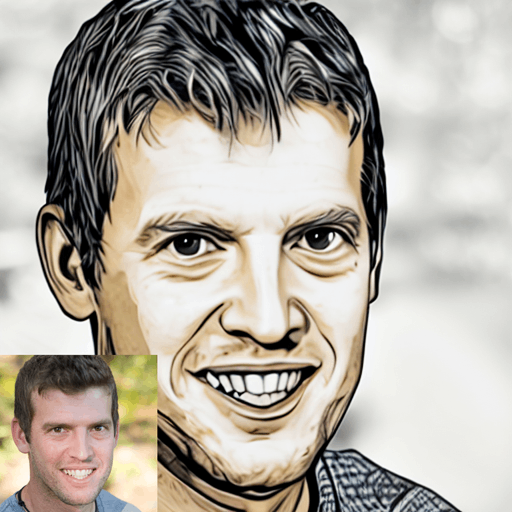}\vspace{1pt}
        \caption{Output 1}
    \end{subfigure}
    \begin{subfigure}[c]{0.115\textwidth}
        \includegraphics[width=1.0\textwidth]{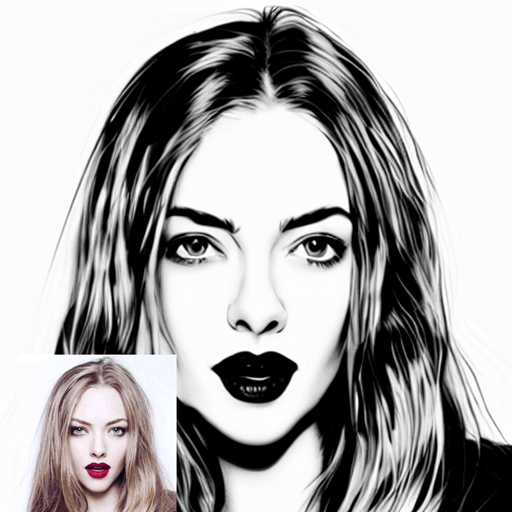}\vspace{1pt}
        \includegraphics[width=1.0\textwidth]
        {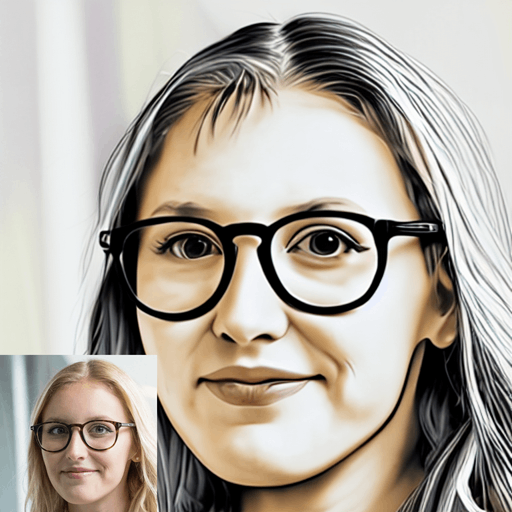}\vspace{1pt}
        \caption{Output 2}
    \end{subfigure}
    \begin{subfigure}[c]{0.115\textwidth}
        \includegraphics[width=1.0\textwidth]{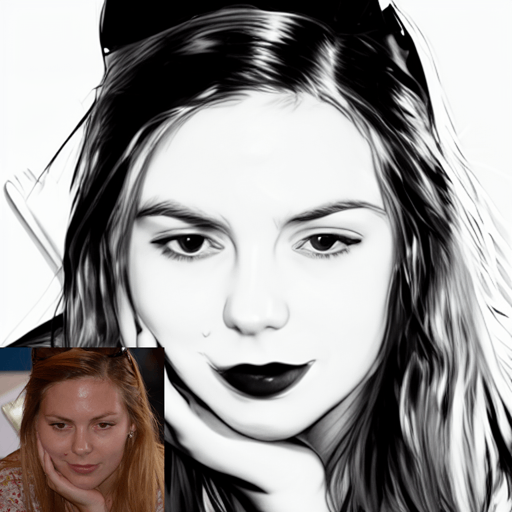}\vspace{1pt}
        \includegraphics[width=1.0\textwidth]
        {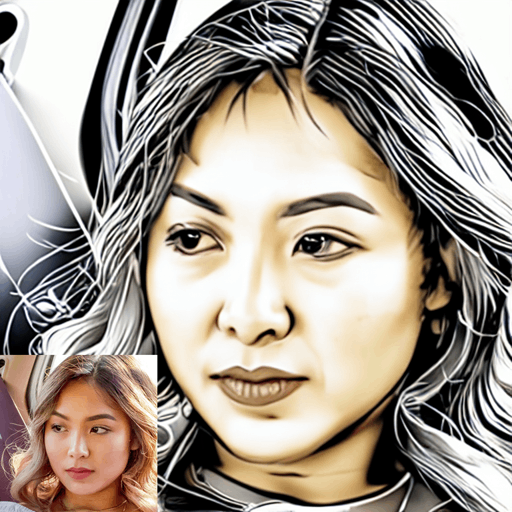}\vspace{1pt}
        \caption{Output 3}
    \end{subfigure}
     \vspace{-2mm}
	\caption{\textbf{Results of transferring real portraits to sketch style.} }
	\label{fig:sketch}
 \vspace{-3mm}
\end{figure}


\vspace{0.5em}
\noindent \textbf{Quantitative Evaluation.} Tables \ref{table:quan_comp} and \ref{table:quan_comp2} present the quantitative comparison results on the CelebAMask-HQ dataset and a mixed dataset from multiple domains, including portraits from CelebAMask-HQ, FFHQ, and AAHQ for a more comprehensive evaluation. As shown, on CelebAMask-HQ, the Gram loss of our method is comparable to the state-of-the-art method \cite{wang2025ppst}, while our LPIPS and ID scores are significantly better, demonstrating superior content and identity preservation. On the mixed dataset, our method achieves the lowest Gram loss. Furthermore, our method outperforms other style transfer techniques in both LPIPS and ID, highlighting its strong domain generalizability.

\vspace{0.5em}
\noindent \textbf{Qualitative Evaluation.}  Figure \ref{fig:comparison-conventional} compares our results with those of previous style transfer methods \cite{wang2025ppst, ye2023ip, zhang2023adding, deng2023zzeroshotstyletransfer, chung2023style}. Our method achieves more faithful stylization effects by effectively transferring styles across semantically aligned regions, such as skin, lips, hair, and background. Additionally, our method demonstrates superior performance in content preservation, unlike artistic methods \cite{ye2023ip, zhang2023adding, deng2023zzeroshotstyletransfer}, which often alter facial identity. Please see the supplementary material for more visual results.

\subsection{More Analysis}
\noindent \textbf{Ablation Studies.} Figure \ref{fig:ablation-ControlNet} investigates the effects of ControlNet and the style adapter. Our high-frequency ControlNet preserves content details more effectively than using a canny edge detection map or the input image. Moreover, with the style adapter, the accuracy of stylization is further enhanced, surpassing the performance of the common IP-Adapter with 4 tokens. Figure \ref{fig:ablation-noise} compares results using different initializations of latent noise. Directly using the input latent as the initial tends to preserve the original color style. Performing AdaIN between the input and reference latents \cite{chung2023style} improves overall stylization effects but doesn't ensure local style similarity. Using the latent of the warped reference yields strong stylization effects, but the output becomes blurry. In comparison, our AdaIN-Wavelet initialization strikes a better balance between stylization and content detail preservation. Quantitative results in Table \ref{table:quan_comp} further support this conclusion. More results can be found in the supplementary material.

\vspace{0.5em}
\noindent \textbf{Controllable Region-specific Style Transfer.} As shown in Figure \ref{fig:regionst}, our method generates natural results by transferring the style of specific semantic regions, such as hair, face, or lips. Given the mask of a specific region, we achieve this by simply replacing the corresponding regions in the warped reference with those from the input portrait. 

\begin{table}[t]
	\centering
    \scalebox{0.90}{
	\begin{tabular}{l|cccc}
		\toprule
		 Metric & Deng et al. &  StyleID & InstantStyle+ & Ours\\
		\midrule
         Time $\downarrow$ (sec) &24.18 & 9.12 & 67.40 & \textbf{6.97} \\ 
		\bottomrule
	\end{tabular}}
    \vspace{-2mm}
     \caption{\textbf{Comparison on inference time.}}\label{table:time}
 \vspace{-3mm}
\end{table}

\vspace{0.5em}
\noindent \textbf{Style interpolation.} Figure \ref{fig:interpolation} demonstrates that our method supports continuous style interpolation, allowing for control over the stylization strength by adjusting the parameter $\gamma$ as described in Section \ref{Initlatent}.

\vspace{0.5em}
\noindent \textbf{Portrait Style Transfer across Different Domains.} Figure \ref{fig:color} demonstrates that our method can colorize grayscale and sketch portraits using reference portraits from different domains. Figure \ref{fig:oldphotos} showcases the ability of our method to restore the color of old photos. Finally, Figure \ref{fig:sketch} illustrates that our method can replicate a sketch style by adjusting the ControlNet conditional scale to 0.4. Please see the supplementary material for more visual results.

\vspace{0.5em}
\noindent \textbf{Inference Time.} Our model processes images at a resolution of 512 $\times$ 512 in approximately 6.97 seconds on an NVIDIA RTX 4090 GPU, making it more efficient than other diffusion-based methods, as shown in Table \ref{table:time}.



\section{Conclusion}
We have presented a portrait style transfer framework that excels in domain generalization and produces high-quality results. By leveraging pre-trained diffusion models and a semantic adapter, we get reliable dense correspondence between portraits, enabling semantically aligned stylization. Besides, we develop an AdaIN-Wavelet transform to balance stylization and content preservation, and also introduce a style adapter as well as a ControlNet to offer style and content guidance. Extensive experiments demonstrate that our method outperforms existing approaches.


\vspace{\baselineskip}
\noindent\textbf{Acknowledgement.} This work was supported by the National Key Research and Development Program of China (2023YFA1008503), the National Natural Science Foundation of China (62471499), and the Guangdong Basic and Applied Basic Research Foundation (2023A1515030002).
 
{
    \small
\bibliographystyle{ieeenat_fullname}

    \bibliography{main}

}


\end{document}